\documentclass[preprint,12pt]{elsarticle}

\usepackage{amssymb}
\usepackage{amsmath,amsfonts}

\usepackage{graphicx}
\usepackage{epstopdf}
\usepackage[dvipsnames]{xcolor}  

\makeatletter
\g@addto@macro\endfrontmatter{\enlargethispage{-2\baselineskip}}
\makeatother

\journal{Elsevier's JVIS}

\begin{document}

\begin{frontmatter}



\title{DRACO-DehazeNet: An Efficient Image Dehazing Network Combining Detail Recovery and a Novel Contrastive Learning Paradigm }


\author[label1,label4]{Gao Yu Lee}

\author[label2]{Tanmoy Dam}

\author[label3]{Md Meftahul Ferdaus}

\author[label1]{Daniel Puiu Poenar}

\author[label4]{Vu Duong}

\affiliation[label1]{organization={School of Electrical and Electronic Engineering, Nanyang Technological University},
            addressline={50 Nanyang avenue}, 
            city={Singapore},
            postcode={639798}, 
            country={Singapore}}

\affiliation[label2]{organization={SAAB-NTU Joint Lab, School of Mechanical and Aerospace Engineering, Nanyang Technological University},
            addressline={50 Nanyang avenue}, 
            city={Singapore},
            postcode={639798}, 
            country={Singapore}}

\affiliation[label3]{organization={Department of Computer Science, The University of New Orleans},
            addressline={2000 Lakeshore Drive}, 
            city={New Orleans},
            postcode={70148}, 
            state={LA},
            country={United States}}
            
\affiliation[label4]{organization={Air Traffic Management Research Institute, School of Mechanical and Aerospace Engineering, Nanyang Technological University},
            addressline={50 Nanyang avenue}, 
            city={Singapore},
            postcode={639798}, 
            country={Singapore}}

\begin{abstract}

Image dehazing is crucial for clarifying images obscured by haze or fog, but current learning-based approaches is dependent on large volumes of training data and hence consumed significant computational power. Additionally, their performance is often inadequate under non-uniform or heavy haze. To address these challenges, we developed the Detail Recovery And Contrastive DehazeNet, which facilitates efficient and effective dehazing via a dense dilated inverted residual block and an attention-based detail recovery network that tailors enhancements to specific dehazed scene contexts. A major innovation is its ability to train effectively with limited data, achieved through a novel quadruplet loss-based contrastive dehazing paradigm. This approach distinctly separates hazy and clear image features while also distinguish lower-quality and higher-quality dehazed images obtained from each sub-modules of our network, thereby refining the dehazing process to a larger extent. Extensive tests on a variety of benchmarked haze datasets demonstrated the superiority of our approach. The code repository for this work is available at https://github.com/GreedYLearner1146/DRACO-DehazeNet.
\end{abstract}




\end{frontmatter}


\section{Introduction}
The presence of haze in images can severely degrade the capability of high-level computer vision algorithms to interpret scenes captured accurately, which greatly affects their performance in image classification, image segmentation, and object detection. This can have catastrophic consequences for mobile-based platforms that rely on vision algorithms, such as autonomous vehicles driving in foggy environments. Therefore, throughout the years, image dehazing has been an important research domain in an attempt to develop effective (and sometimes efficient) dehazing algorithms to complement subsequent computer vision tasks. 

Current State-Of-The-Art (SOTA) dehazing approaches rely on designing deep neural network architectures or, more recently, Vision Transformers (ViTs). Such dehazing algorithms are trained in an end-to-end manner on benchmark haze datasets, which often involve paired hazy and ground-truth images. This implies that their performances is heavily dependent on the amount of data available. Many dehazing works have utilized dehazing benchmark based on artificial haze generation techniques on image datasets with corresponding depth maps (e.g., NYU2 \cite{silberman2012indoor}, RESIDE \cite{li2018benchmarking}), which usually involved the Koschmieder model, a subset of a class of Atmospheric Scattering Models (ASM) \cite{koschmieder1925theorie}. Such datasets are typically of huge quantity (RESIDE comprises 110,500 synthetic hazy images and only 4807 real hazy images). However, real-life hazy images are relatively difficult to obtain (examples include I-HAZE \cite{ancuti2018haze}, O-HAZE \cite{ancuti2018ohaze}, NH-HAZE \cite{ancuti2020nh}, DENSE-HAZE \cite{ancuti2019dense} and RS-HAZE \cite{song2023vision}), and are of lower quantity (O-HAZE only comprises 45 hazy and clear images). Although extensive studies have been devoted to evaluating the proposed algorithm on the latter's class of haze datasets, we found that most of the approaches emphasized on improving dehazing performance by significantly increasing the width or depth of the algorithmic architecture, leading to high computational demand and rendering their usage unsuitable for platforms with limited computational resources, such as autonomous cars and Unmanned Aerial Vehicles (UAVs). In these classes of approaches, a number of them (e.g., Zhang et al. \cite{zhang2020drcdn}, Feng et al. \cite{feng2021urnet} and Yang et al. \cite{yang2022transformer}) proposed the Residual Network (ResNet) blocks, which can alleviate the vanishing gradient problem but could lead to overfitting and higher computational demands in smaller datasets. 

Most studies addressed the dehazing efficiency by either reducing the number of widths or depths of the algorithm, or by applying Knowledge Distillation (KD) (e.g., Lan et al. \cite{lan2022online}, Hong et al. \cite{hong2020distilling} and Wang et al. \cite{wang2022multi}). Some studies (e.g., Feng et al. \cite{feng2019image}, Deivalakshmi et al. \cite{deivalakshmi2022deep} and Kuanar et al. \cite{kuanar2019night}) complement the aforementioned solutions via dilated convolution instead of ordinary convolution, which has the advantage of expanding the receptive field of the network without sacrificing resolution. However, for the former, determining the optimal trade-off between the width and depth of the dehazing architecture is a challenging task that can only be achieved via extensive ablation studies. As for the latter, the computational costs depend on how architecturally complex the teacher model is, since the internal features extracted from the teacher model to be transferred to the student model are performed iteratively. In the classification domain, apart from implementing either of the two  aforementioned approaches, there exist widely agreed architecture that promise effective and efficient task performance, such as MobileNetV1 \cite{howard2017mobilenets}, MobileNetV2 \cite{sandler2018mobilenetv2}, and EfficientNet \cite{tan2019efficientnet}, all of which utilize the inverted residual block instead of the ResNet blocks. However, no such framework has been widely agreed upon for dehazing, and it would be interesting to explore how, and whether incorporating inverted residual blocks could contribute to effective and efficient dehazing. 

Another issue is that the role of detail recovery in the development of dehazing algorithm has not been widely studied. This is preliminary pointed out by Li et al. \cite{li2022single}, and has only been applied in a few dehazing works as of writing of this work (e.g., the aforementioned work \cite{li2022single} and Fang et al. \cite{fang2022detail}). Image deraining, a related domain of image restoration, see more works incorporating detail recovery (e.g., Gao et al. \cite{gao2022heavy}, Deng et al. \cite{deng2020detail}, Shen et al. \cite{shen2022detail} and Zhu et al. \cite{zhu2022hdrd}). Because detail loss may be inevitable during a typical dehazing procedure, a separate detail recovery network helps to restore the intrinsic detail feature map, which is then fused with the originally coarse dehazed images to recover and enhance the outputs. Taking detail recovery into account has demonstrated improvements for medium to heavy blur in deraining (as illustrated by the aforementioned \cite{gao2022heavy}, Ahn et al. \cite{ahn2022remove} and Jiang et al. \cite{jiang2023two}). To the best of our knowledge, there are few approaches that focus on detail recovery for non-homogeneous and thick real haze scenarios (as in NH-HAZE and DENSE-HAZE). Given that the algorithms evaluated on the latter two datasets gave sub-par performances relative to the other homogeneous haze datasets, it is also of interest to explore how the dehazing performance may fare under heterogeneous and thick haze conditions when detail recovery is invoked in a dehazing architecture.

Most proposed dehazing models fall under prior-based or learning-based approaches. For the latter, CNN-based dehazing models are the common norm, until the advent of ViT-based dehazing approaches (e.g., Song et al. \cite{song2023vision}, Zhao et al. \cite{zhao2021complementary}).  More recently, the contrastive learning paradigm, initially utilized for classification (e.g., triplet network \cite{hoffer2015deep}, ProtoNet \cite{snell2017prototypical}), has been modified for dehazing. The latter class of approaches has been promising in addressing the challenges of achieving dehazing effectiveness under real hazy data of small sample sizes (for empirical results and trends in this domain see Lee et al. \cite{lee2024dehazing}). For example, Wu et al. \cite{wu2021contrastive} introduced a contrastive regularization dehazing network (AECR-Net) that utilized a triplet network-based contrastive learning architecture. The dehazed, ground-truth and hazy images served as the anchor, positive and negative inputs respectively, and the regularization procedure constrained the anchor images such that the extracted features were clustered closer to those of the positive images, and pulled away from those of the negative images. This allowed effective dehazing to be achieved without requiring additional parameters and computational demands during the inference phase, as the contrastive module refines the intermediate dehazing output and can be removed during inference. Their approaches also empirically demonstrated, for the first time (to the best of our knowledge), that contrastive learning could be extended to other visual tasks apart from classification. However, the triplet network-inspired architecture of the AECR-net implies that they only take one particular set of negative samples, and this means that the refinement process may be limited as there are fewer diverse cues to be utilized for better clustering of the features. More recent works such as C$^{2}$PNet \cite{zheng2023curricular} and CARL \cite{cheng2022robust} have explored the role of increasing negative samples to add information and hence versatility for the contrastive paradigm, which has indeed demonstrated further enhancement of the intermediate dehazed outputs. However, for the two aforementioned methods, outputs from other existing dehazing models are required, which adds algorithmic complexity to the respective method, and the results obtained may not be consistent for the different dehazing models utilized.

Therefore, motivated by the need for a more efficient and effective dehazing model that accounts for detail recovery under data size constraints, we proposed a Detail Recovery And COntrastive learning-based Dehaze Network (DRACO-DehazeNet), which was inspired by DPE-Net by \cite{gao2022heavy} for image deraining. Our network not only proposed the Dense Dilated Inverted Residual Block (DDIRB), which is a more efficient version of the Dense Dilated Residual Block (DDRB) proposed in DPE-Net, but also includes an ATTention-imbued Detail Recovery architecture (ATTDRN) that reduces the dehazing artifacts after DDIRB and enhances its coarse output. The ATTDRN takes on a similar role as the Enhanced Residual Pixel Attention Block (ERPAB) in DPE-Net, but combines both channel and spatial attention in a densely connected manner. Finally, the contrastive dehazing procedure is performed jointly with the DDIRB and ATTDRN using a modified quadruplet training network approach (the original framework is inspired from Chen et al. \cite{chen2017beyond} for person re-identification), which entailed the extracted features of the intermediate dehazed outputs from the DDIRB and ATTDRN, as well as the hazy and clear images, thereby diversifying the negative samples in the training loop via the addition of the DDIRB outputs. The upshot of our network is that it not only re-emphasizes the role and feasibility of contrastive learning and detail recovery on dehazing tasks, as well as enhancing the intermediate dehazing outputs via a quadruplet loss-based paradigm, but also pioneers the use of inverted residual blocks to reduce the computational demands during haze removal. 

We realized that there is actually one other recent dehazing work as of writing that utilizes the quadruplet loss (i.e., LIDN by Ali et al. \cite{ali2023lidn}). Their proposed network agreed with our argument on the usage of quadruplet networks to reduce haze artifacts and further refining the dehazing outputs. However, their network implemented four sub-modules, each of which culminates into one loss function, and the four loss functions are known as the ``quadruplet loss", which is not the type of loss utilized in the contrastive learning framework involving computing and clustering the distances between the embedded extracted features from each of the image set.


    

To the best of our knowledge, our network uniquely combines dilated inverted residual networks, detail recovery and quadruplet network-based contrastive learning into one dehazing architecture tested on both synthetic and real-world haze datasets of varying uniformity, density and size. Specifically:

\begin{itemize}

\item We pioneered the use of efficient inverted residual blocks over commonly utilized residuals to enable lighter yet more accurate dehazing models, which are feasible even on mobile platforms. Although we adopted a two-stage design with separate DDIRB and ATTDRN components, our quadruplet network-based contrastive procedure innovatively complements them to collectively push dehazing accuracy to new levels, outperforming state-of-the-art techniques even with limited training data.

\item Extensive experiments on smaller, real-world haze benchmarks such as O-HAZE, NH-HAZE and DENSE-HAZE, as well as larger synthetic datasets such as RESIDE, demonstrate the superiority of our design in terms of the widely utilized effectiveness metrics such as PSNR and SSIM alongside efficiency gains in computational overhead measured by FLOPs. We consistently show strong generalization and accuracy across diverse haze types while requiring fewer computational resources.
\end{itemize}

Our proposed DRACO-DehazeNet sets new standards for effectively handling varying haze environments under tight data and computational constraints via its unique fusion of dilated inverted residuals, detail recovery, and quadruplet network-based contrastive learning. The potential domain applications ranging from autonomous navigation to aerial imaging are also significantly advanced through our more feasible yet performant architecture.

\section{Related Works}
Image dehazing has been a major focus of research for over ten years. This field has a diverse range of literature exploring various algorithms to restore visuals affected by haze. In this section, we provide background on the image formation model related to haze and review key developments in dehazing techniques. We look at early approaches based on set principles, the shift to data-driven methods using deep learning, and the growing interest in vision transformers. We also consider the limitations of the current literature, which leads to the new techniques we propose. This section prepares the reader for the innovations we discuss later.

First, we describe a common model for image degradation due to atmospheric scattering, which helps us understand how haze forms. We then review important research that uses established rules and assumptions to estimate transmission maps, which are crucial for dehazing a single image. Subsequently, we explain how deep convolutional and transformer networks have been developed to predict haze-free scenes directly from the data. Among these learning-based methods, we focus on the use of attention mechanisms and multi-scale processing as significant advancements. Finally, we address major issues such as dependence on large datasets, computational inefficiency, and reduced effectiveness in dense haze. Our proposed solution, DRACO-DehazeNet, aims to address these challenges by using a unique combination of dilated inverted residuals, detail recovery, and contrastive regularization. The literature review not only explains fundamental concepts but also highlights where our methodological contributions fit.

\subsection{The Atmospheric Scattering Model}

As previously mentioned, the Atmospheric Scattering Model (ASM) is a common mathematical model for modelling and generating haze. The ASM is mathematically described as

\begin{equation}
    I(x) = J(x)t(x) + A(1-t(x)),
\label{eq1}
\end{equation}

where $I(x)$ and $J(x)$ represent the hazy and clear images respectively ($x$ denotes the pixel coordinates), $A$ is the atmospheric light value contributed by environmental illumination, and $t(x)$ is the transmission map related to the depth of the scene via $t(x) = e^{-\beta d(x)}$, where $\beta$ is the atmospheric scattering coefficient. The term $J(x)t(x)$ is also known as the direct attenuation $D(x)$ and the term $A(1-t(x))$ is also known as the airlight $L(x)$ \cite{ancuti2020day}. Hence, Equation \ref{eq1} can also be written as the sum of the two contributions:

\begin{equation}
    I(x) = D(x) + L(x).
\label{eq2}
\end{equation}

The airlight is the result of both sunlight infused with haze particles and light scattered from these particles, leading to contrast distortion and reduction. In haze datasets that involve generating artificial haze using equation \ref{eq1}, $\beta$ is an essential factor that determines the intensity or thickness of the haze generated in the desired context.

\subsection{Inverted Residual Block vs Residual Block}

\begin{figure}
\centering
\includegraphics[scale=1.15]{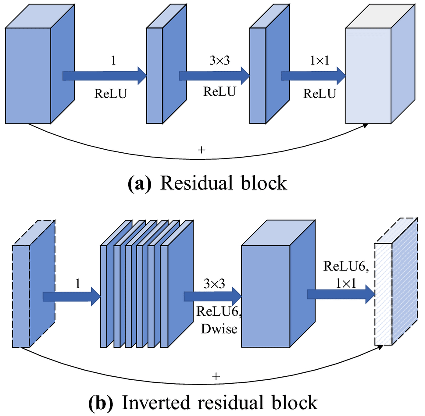}
\caption{Comparison of the ordinary ResNet block (\textbf{top}) with the inverted ResNet block (\textbf{bottom}). The image is adapted from Ding et al. \cite{ding2022slimyolov4}.}\label{fig2.01}
\end{figure}

A schematic diagram of an ordinary residual block and the inverted residual block is illustrated in Figure \ref{fig2.01}. The original residual block utilized a ``wide $\rightarrow$ narrow $\rightarrow$ wide'' structure which entailed the widening of the input (usually of a high channel) via a 1 $\times$ 1 convolution operation. The intermediate outputs are then compressed (narrow) via a 3 $\times$ 3 convolution operation before being widened again via another 1 $\times$ 1 convolution.

On the contrary, the inverted residual block utilized a ``narrow $\rightarrow$ wide $\rightarrow$ narrow'' structure which is the opposite of that of the residual block, hence its name. Firstly, a 1 $\times$ 1 convolution is used to narrow the input feature, followed by widening the intermediate features via a 3 $\times$ 3 depth-wise convolution operation (instead of the ordinary convolution). Lastly, another 1 $\times$ 1 convolution is used to narrow the features again. By utilizing the depth-wise convolution in the middle of the inverted residual block structure, the number of training parameters required can be greatly reduced as the image transformation are only performed once before elongated into $N$ channels, hence removing a bulk of the tedious image processing and transformation (unlike the ordinary convolution which involved the transformation of the image $N$ times). This also reduces the FLOPs and computational memory incurred.

Both the ordinary and inverted residual block utilized skip connection  to connect its initial and the final layer so as to enhance the gradient flow along the blocks. However, since the inverted residual block entailed widening the middle layer of its structure through a larger kernel size, and that widened architecture of such type has been shown to reduce gradient confusion, the performance of the model can be enhanced greatly and trained at a faster rate \cite{sankararaman2020impact}.

\subsection{Prior-based Dehazing}
Some of the first dehazing algorithms proposed rely on understanding certain priors or contexts of hazy images to estimate dehazing parameters, such as transmission maps, to guide the dehazing process. These are known as prior-based dehazing approaches, for which Dark Channel Prior (DCP) by He et al. \cite{he2010single} is one of the best-known approaches. This prior assumed that haze-free images contain dark pixels, region(s) not covering the sky with a minimum of one RGB color channel of lower pixel intensity. In hazy images, the ambient light $A$ contributes mostly to these dark pixels and is utilized for direct transmission map estimation to recover the dehazed images. The DCP was one of the first approaches to tackle single image-based dehazing. There are other prior-based dehazing methods that utilize different assumptions on hazy images. Fattal \cite{fattal2008single} utilized the prior on the surface shading function being statistically uncorrelated with the scene transmission over a local set of pixels in haze images for dehazing. It does not incorporate the ASM, and hence may introduce color artifacts and contrasts. The Colour Attenuation Prior (CAP) were introduced by Zhu et al. \cite{zhu2015fast}, and the transmission depth was estimated by assuming that the saturation and brightness of a pixel in a hazy scene can be modelled linearly with respect to the depth. However, CAP is not sensitive to objects or scenery with inherent white color intensities, and may lead to inaccurate estimates in these regions.

A common limitation in utilizing prior-based dehazing approaches is that their effectiveness is largely dependent on the image scene context and haze distribution. Therefore, a prior method that is effective for one scene may not be effective for other prior methods, and vice versa. Another limitation is that they often perform poorly in non-homogeneous and thick haze scenarios, as demonstrated by numerous studies (e.g., AECR-Net \cite{wu2021contrastive} and Wei et al. \cite{wei2021non}) using DCP.

\subsection{Learning-based Dehazing}
Learning-based approaches deploy deep learning techniques to directly learn the haze parameters for dehazing. Some learning-based models invoked ASM to guide the estimation after the learning process, whereas others directly establish the mapping between hazy and clear images. As mentioned in the introduction, because ViT-based approaches have been recently proposed for dehazing, we will discuss CNN-based and ViT-based dehazing approaches in separate subsections.

\subsubsection{CNN-based Dehazing}

DehazeNet by Cai et al. \cite{cai2016dehazenet} is one of the earliest deep learning CNN-based dehazing approaches that performed dehazing using only hazy-clear corresponding image pairs under the guidance of the ASM. The Multi-Scale CNN (MSCNN) dehazing by Ren et al. \cite{ren2016single} involved a coarse net architecture that learned the transmission map of the hazy images as a whole, and a fine net architecture that refined the dehazed outputs in a local manner while using the ASM. Although both DehazeNet and MSCNN yielded visually pleasing outputs, the dehazing performances was still limited as the ambient light was still estimated via ASM. The All-in-One Dehazing Network (AOD-Net) was proposed by Li et al. \cite{li2017aod} and addressed the aforementioned issue via a lightweight architecture that combined the ambient light and the transmission map into a single parameter for more effective estimation and restoration of the dehazed images. An attention-based feature fusion mechanism was incorporated into the Feature-Fusion Attention Network (FFA-Net), proposed by Qin et al. \cite{qin2020ffa}. It incorporates channel-wise and pixel-wise attention to address the different features in a weighted manner to extend the representational capability of the CNN and its flexibility to handle different types of information. GridDehazeNet by Liu et al. \cite{liu2019griddehazenet} introduced multi-scale estimation on a grid network via an attention mechanism that mitigates the bottleneck issue commonly encountered in typical multi-scale approaches. 


There are also dehazing methods that take advantage of dense connections to improve information flow along features from different level and further reduce the vanishing gradient problem during learning. For instance, Zhang et al. \cite{zhang2018densely} introduced the Densely Connected Pyramid Dehazing Network (DCPDN), which jointly optimized the transmission map, atmospheric light estimation and image dehazing tasks by leveraging features from different levels via pooling modules. Wang et al. \cite{wang2022msf} proposed a Multi-Scale Feature Fusion Dehazing Network with Dense Connection ($MSF^{2}DN$) to combine features extracted from different convolutional layers repeatedly before being fed to the respective feature fusion modules. A multiple inter-scale dense skip connection dehazing network (MSNet) was proposed by Yi et al. \cite{yi2021msnet} which considers complimentary information via dense inter-scale skip-connections in the encoder and decoder, as well as a bottleneck residual block layer that manipulates the weights of the local gradients at different scales. All of the aforementioned works performed exceptionally well on synthetic and real-world datasets with approximately homogeneous haze, although only the $MSF^{2}DN$ was tested on thick and non-homogeneous haze scenarios. In addition, a few aforementioned works have also been evaluated in the non-homogeneous and thick haze scenario (e.g., DehazeNet, AOD-Net, FFA-Net, GridDehazeNet), and all have shown consistently better performances relative to prior-based approaches, although the visual outputs still have vast room for improvements.

CNN-based approaches commonly outperform prior-based approaches in many different contexts and datasets, including those discussed in the introduction. However, their performance is strongly dependent on the amount of available training data. As mentioned earlier, the aforementioned works do not explore the role of contrastive learning, inverted residual blocks and detail recovery in the dehazing architecture. Even methods utilizing the detail recovery, which is highlighted in the introductory section, did not explore their methods on the non-homogeneous and thick haze scenarios.

\subsubsection{ViT-based Dehazing}

As mentioned in the introductory section, Song et al. \cite{song2023vision} proposed the DehazeFormer which is inspired by the shifted window (swin) ViT architecture, but with several key modifications, such as a modified normalization layer, activation function, and a spatial information aggregation scheme. The modifications are performed as in the original swin ViT architecture, and the modules were not very effective for image dehazing. Several variants of the model are introduced and compared: DehazeFormer-T, DehazeFormer-S (required 1.283M parameters), DehazeFormer-B (required 2.514M parameters), DehazeFormer-M (required 4.634M parameters), and DehazeFormer-L. A Hybrid Local-Global (HyLoG) attention-based architecture was introduced by Zhao et al. \cite{zhao2021complementary}, which incorporated both local and global transformer paths to capture global and local feature dependencies in the images simultaneously. This was integrated into their Complementary Feature Selection Module (CFSM) to adaptively select essential complementary features in dehazing tasks. Dehamer was proposed by Guo et al. \cite{guo2022image} and is one of the first works to combine a CNN with ViT in dehazing. To handle feature inconsistency between CNN and transformer modules, they proposed a feature modulation procedure on CNN-captured features, which allows their procedure to inherit both the global and local feature modelling capabilities of the transformer and CNN, respectively, in a simultaneous manner. 

ViT approaches typically utilize more parameters owning to the more complex architecture of some of the proposed designs (e.g., HyLoG-ViT). Although they outperformed CNN approaches on the same benchmark datasets and settings, they consumed more computational resources; hence, it might be challenging to implement them for small dataset, as well as for efficient real-time mobile platform operations. A few aforementioned ViT works (e.g., Dehamer) have been evaluated in the non-homogeneous and dense haze dataset and yielded better metrics values than a majority of the described CNN-based works. However, the dehazed outputs still leave room for improvement. Finally, like CNN-based approaches, transformer-based works also did not explore the role of contrastive learning and detail recovery in the dehazing architecture.

\subsubsection{Effect of Dilated Convolution Rates on Feature Extraction}

\begin{figure}[hbt!]
    \centering
    \includegraphics[scale=0.50]{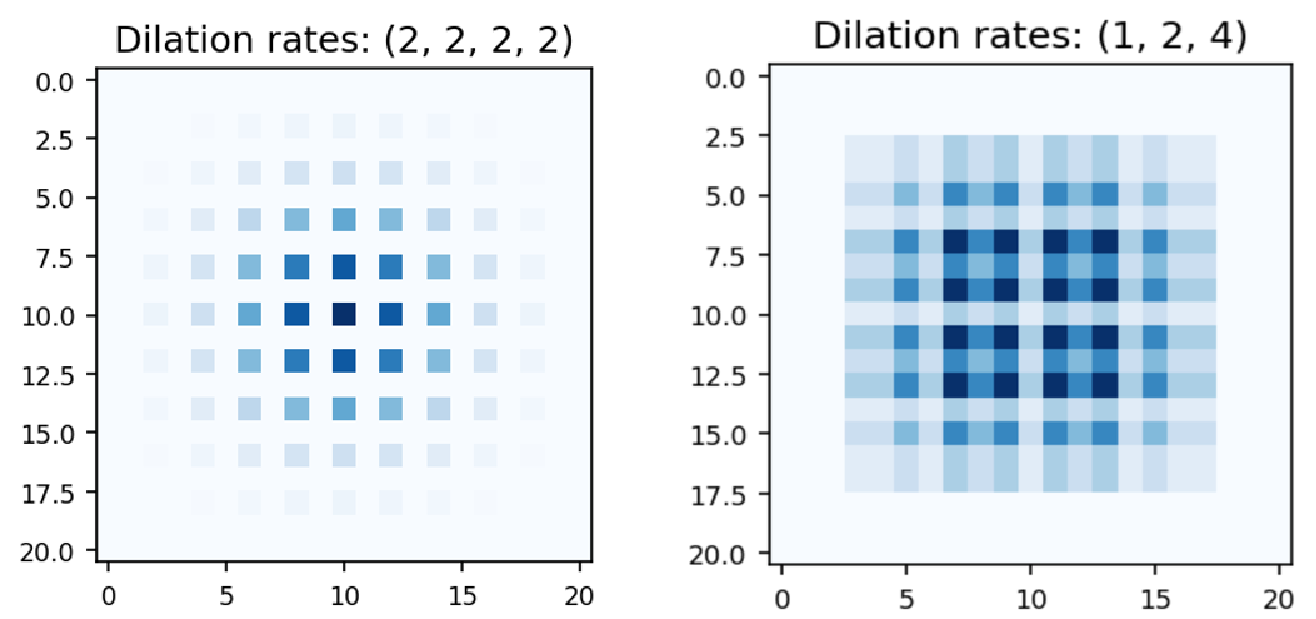}
    \caption{Illustration of the gridding problem that arose when using the same vs different dilation rates. The purpose of the dilated convolution is to output a big receptive field encompassing the entire image for a given center pixel, as shown on the left of the figure. If we utilize the same dilation rate (2) throughout, we can also see a uniform grid of pixels which does not contribute to the output center pixel, illustrating the unwanted gridding problem. Conversely, if we were to use differing dilation rate (1,2,4) as in the right of the figure, pixels that are closer to the centre pixel to contribute even more than that of those further away, hence reducing the checkerboard receptive field issue. Thefigure is adapted from the github page ``Dilation Rate Gridding Problem and How to Solve It With the Fibonacci Sequence" (https://github.com/Jonas1312/dilation-rate-as-fibonacci-sequence)}
    \label{gridding}
\end{figure}

The aforementioned CNN-based works mostly did not take into account the role of dilated convolution, even for those that involved multiple network branches. Dilated convolution served to increase the efficiency of a CNN model via widening the receptive kernel size so that the amount of convolution computation for a fix image size would be lesser. They also allowed feature extraction at multiple scales, especially if different dilation rates are utilized consecutively. Although the model is able to compute broader image contexts, some information loss (especially the spatial resolution) would be inevitable due to the fact that the kernel size now has spacing between their pixel window, leading to pixel skipping and hence a ``checkerboard" perspective as highlighted by Wang et al. \cite{wang2018understanding}. With a larger dilation rate, more spatial resolution loss would be incurred as the feature extracted would be spread over a wider kernel window. Dilated convolution has been used in works like DDRH-Net by feng et al. \cite{feng2019image} and LKD-Net by Luo et al. \cite{luo2023lkd}, but such models utilized a set of fixed dilation values (2 for DDRH-Net and 3 for LKD-net). When different dilation rates were used, the gridding problem can be mitigated, and has been demonstrated by Wang et al. \cite{wang2018understanding}. This is especially so for the dilation rate combination of 1, 2 and 4. The gridding problem is best explained by illustration, which is shown in Figure \ref{gridding}.

\section{Our approach}

In this section, we describe in detail our proposed DRACO-DehazeNet architecture, as well as elaborate on the respective modules incorporated in our design, specifically DDIRB, ATTDRN, and the contrastive quadruplet network. 

\begin{figure}[hbt!]
    \centering
    \includegraphics[scale=0.69]{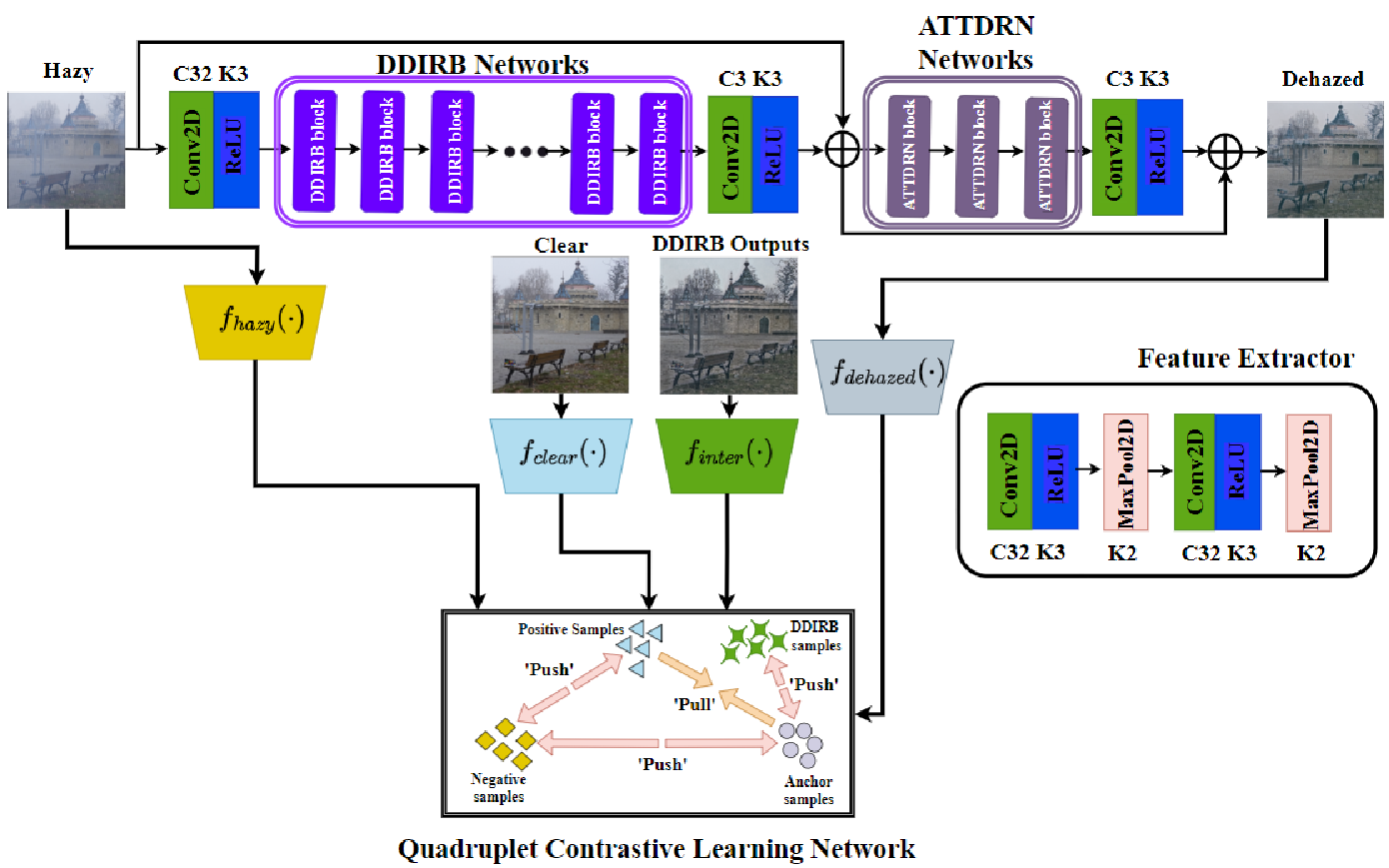}
    \caption{Illustration of the overall architecture of our DRACO-DehazeNet. Once again, C means the number of channels, K means the kernel size, and D means the dilation rate. All strides used is of value 1. The component of the feature extraction module are also illustrated for completeness.}
    \label{DRACO_DehazeNet2}
\end{figure}

\subsection{Overall DRACO-DehazeNet}

The overall structure of our DRACO-DehazeNet is depicted in Figure \ref{DRACO_DehazeNet2}. It comprised of a 2D convolutional layer followed by sequential blocks of DDIRB module (the DDIRB Network), for which the number of blocks selected is 3. The intermediate outputs is fed to another 2D convolutional layer and added with the original input before passing through the ATTDRN network. The number of ATTDRN blocks in the latter network is also set as 3, and the corresponding outputs are fed to yet again a 2D convolutional layer before being added with the intermediate output before the ATTDRN operation. In all convolutional layer, the ReLU activation function was utilized throughout, with the respective kernel and channel size specified in Figure \ref{DRACO_DehazeNet2}, all with a stride of 1.

The original hazy images, the intermediate outputs from the DDIRB, the dehazed images from the DDIRB and ATTDRN, and the clear images have their feature extracted as part of our quadruplet network contrastive learning process. The feature extractor comprised of two 2D convolutional layer and two max pooling layer, as illustrated in Figure \ref{DRACO_DehazeNet2}.  The quadruplet contrastive learning involved clustering the features from the intermediate DDIRB outputs and the ground-truth clear images, and repelling the features from the hazy images and clear images, as well as between the dehazed features and DDIRB outputs and hazy image features. This procedure is also shown in the same figure.


\subsubsection{DDIRB}

The DDIRB architecture illustrated in Figure \ref{DRACO_DehazeNet1} is a densely connected version of the inverted residual block introduced in EfficientNet \cite{tan2019efficientnet} for image classification. Dense connection ensures that the vanishing gradient problem can be alleviated greatly, thus serving as one way to enhance our model's dehazing capability. Each block comprised an ordinary 2D convolutional layer with a kernel and stride size of 1, followed by a depth-wise convolutional layer with a kernel size of 3 and stride size of 1. The latter layer ensures computational efficiency by convolving the images with each of the respective channels (or depths) separately and stacking them back after all channels have been convolved (instead of performing such tasks over all channels simultaneously). A Squeeze and Excite (SE) Network (also illustrated in Figure \ref{DRACO_DehazeNet2}) \cite{hu2018squeeze} is then applied, which improves the channel inter-dependency by including a content-aware mechanism for adaptive weight assignment to each channel (instead of assigning the same weights) in the previous depth-wise layers. The features were then passed through another ordinary 2D convolutional layer with the same configuration before they were finally added to the original input features. The algorithmic operations are then repeated for the subsequent two blocks, with the features of each consecutive block densely connected with those from all possible previous blocks, with an increasing dilation rate of 2 for the second block and 5 for the third block. These dilation rates were selected to handle the gridding problem as already pointed out earlier. All the activation function utilized were ReLU functions. The parameter settings for each component of the single DDIRB block are listed in Table 1.

\begin{table} 
\centering
\caption{The parameter setting for a single DDIRB block in our DRACO-DehazeNet. The number on the left of each row of `Layers' denotes the layer number of the DDIRB block.}
\begin{tabular}{p{4cm}p{2cm}p{2cm}p{2cm}p{2cm}}
\hline
\textbf{Layers} & \textbf{Kernel} & \textbf{Dilation} & \textbf{Input Channel} & \textbf{Output Channel}\\
\hline 
 (1) Conv2D & 1$\times$1 & 1 & 32 & 32\\
 (2) Depthwise & 3$\times$3 & 1 & 32 & 32\\
 (3) SE block  & - & 1 & 32 & 32\\
 (4) Conv2D & 1$\times$1 & 1 & 32 & 32\\
 (5) Add (1)+(4) & - & - & 32 & 32 \\
 (6) Conv2D & 1$\times$1 & 2 & 32 & 32\\
 (7) Depthwise & 3$\times$3 & 2 & 32 & 32\\
 (8) SE block & - & 2 & 32 & 32\\
 (9) Conv2D & 1$\times$1 & 2 & 32 & 32\\
 (10) Add (5)+(9) & - & - & 32 & 32 \\
 (11) Conv2D  & 1$\times$1 & 5 & 32 & 32\\
 (12) Depthwise & 3$\times$3 & 5 & 32 & 32\\
 (13) SE block  & - & 5 & 32 & 32\\
 (14) Conv2D & 1$\times$1 & 5 & 32 & 32\\
 (15) Add (10)+(14)  & - & - & 32 & 32 \\
\hline
\end{tabular}
\label{tab:perturbation_classification}
\end{table}

The effect of the overall DDIRB structure is to remove as much haze as possible from a given image context, and thus the architecture encourages the usage of sequential inverted residual block (and hence convolution blocks), culminating in a deep neural network structure. Despite utilizing such structure, the efficiency of the network is greatly reduced via the depth-wise convolution layers imbued in between the ordinary convolutional layers for each inverted residual block, as already highlighted earlier.

\subsubsection{ATTDRN}

\begin{figure}[hbt!]
    \centering
    \includegraphics[scale=0.60]{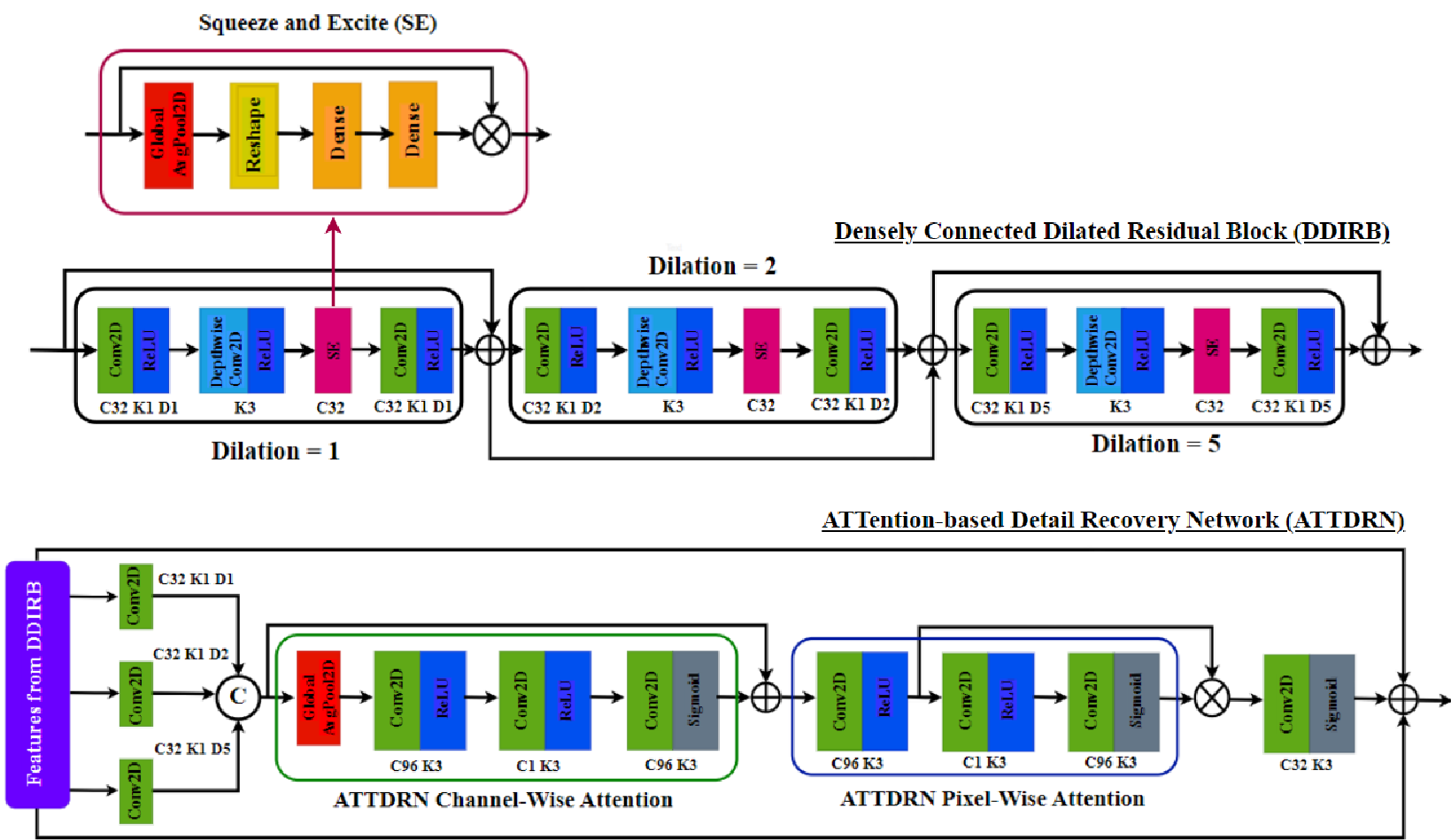}
    \caption{Illustration of the Densely Connected Dilated Residual Block (DDIRB), along with the ATTention-based Detail Recovery Network (ATTDRN) component of our DRACO-DehazeNet. The Squeeze and Excite (SE) block component is also illustrated. In each individual block of the components, C represents the number of channels, K represents the kernel size, and D represents the dilation rate. All strides used is of value 1.}
    \label{DRACO_DehazeNet1}
\end{figure}

The ATTDRN architecture utilized both channel and pixel attention unlike the ERPAB in DPE-Net, which addressed the observations that most dehazing networks handle the channel and pixel-wise feature components equally, and thus is not adaptive enough to deal with non-homogeneous haze and weighted channel-wise features. The ATTDRN architecture is also illustrated in Figure \ref{DRACO_DehazeNet1} and is a combination of the first few components of the ERPAB and a densely connected version of the FFA-net attention modules. First, the intermediate features from the DDIRB output were passed through three convolutional layers with the same kernel, strides and filter sizes but different dilation rates (1, 3 and 5) in a parallel manner before concatenation (and once again dealing with the gridding problem). The concatenated features are then passed through a channel-wise attention module comprising of an average pooling layer, three convolutional layers, and an addition layer that sums the feature tensor from the convolutional and concatenated layers. The obtained features are then inserted into a pixel-wise attention module comprising the same components in the channel attention module except for average pooling. The final layers involved the addition of the pixel attention module features directly to the DDIRB-extracted features. Similar to the DDIRB, all stride values utilized were 1. Once again, all the activation functions utilized are the ReLU function, and the parameter settings for each component of a single ATTDRN block are outlined in Table 2. 

\begin{table} 
\centering
\caption{The parameter setting for a single ATTDRN block in our DRACO-DehazeNet. The number on the left of each row of `Layers' denotes the layer number of the ATTRDN block.}
\begin{tabular}{p{4cm}p{2cm}p{2cm}p{2cm}p{2cm}}
\hline
\textbf{Layers} & \textbf{Kernel} & \textbf{Dilation} & \textbf{Input Channel} & \textbf{Output Channel}\\
\hline 
(1) Conv2D  & 3$\times$3 & 1 & 1 & 32 \\
(2) Conv2D  & 3$\times$3 & 1 & 2 & 32\\
(3) Conv2D  & 3$\times$3 & 1 & 5 & 32\\
(4) Concat (1)-(3) & - & - & - & 96\\
(5) Avg Pooling &  & 1 & 96 & 96\\
(6) Conv2D  & 3$\times$3 & 1 & 96 & 96 \\
(7) Conv2D  & 3$\times$3 & 1 & 96 & 1\\
(8) Conv2D  & 3$\times$3 & 1 & 1 & 96\\
(10) Add (4)+(8) & - & - & 96 & 96\\
(11) Conv2D  & 3$\times$3 & 1 & 96 & 96 \\
(12) Conv2D  & 3$\times$3 & 1 & 96 & 1\\
(13) Conv2D  & 3$\times$3 & 1 & 1 & 96 \\
(14) (11)$\times$(13) & - & - & 96 & 96\\
(15) Conv2D & 3$\times$3 & 1 & 96 & 32\\
(16) Add input+(15)  & - & - & 32 & 32\\
\hline
\end{tabular}
\label{tab:perturbation_classification}
\end{table}

The effect of the overall ATTDRN structure is the maximal removal of dehazing artifacts left behind from DDIRB via a combined channel-spatial attention modules in sequence, since such a mechanism selectively highlights regions containing the remaining artifacts. The concatenation of the different intermediate outputs of different dilated rates at the beginning of the module functioned as a multiscale dehazing artifacts analyzer, further complementing the detail recovery process. However, as we shall see in the ablation study section, merely utilizing the ATTDRN module in our model would result in image artifact like color deviation since it is designed primarily for post-dehazing processing.

\subsubsection{Contrastive Quadruplet Network}

There are two aspects to be considered in the contrastive learning procedure following Wu et al. \cite{wu2021contrastive}. The first is to generate positive and negative pairs, and the second is to compute the latent feature space of these pairs for the learning procedure. The first aspect is addressed by considering that the positive pairs comprise the anchored (dehazed) images obtained from the DDIRB-ATTDRN modules and the ground-truth clear images, whereas the negative pairs comprise the same dehazed images and the hazy images. The second aspect was addressed by utilizing the VGG19 architecture \cite{simonyan2014very} as the feature extraction module of the images from the positive and negative pairs. 

Deviating from the triplet network paradigm, a second negative pair was created via intermediate outputs from the DDIRB block. This is based on the prior assumption that the dehazed images yielded without detail recovery are of lower image quality than the corresponding images with detail recovery. As emphasized in the Detail Recovery Network (DRN) approach by Li et al. \cite{li2022single}, this is because the information from the image could be inevitably lost during the dehazing process, and halo artifacts could be resulted due to lack of information about that particular image region in its transmission map computation during the dehazing process. The MSCNN approach supported this hypothesis in that the scene transmission map without the detail recovery are estimated more poorly than that with the latter, thus affecting the subsequent quality of the dehazed images.

\begin{figure}[hbt!]
    \centering
    \includegraphics[scale=0.70]{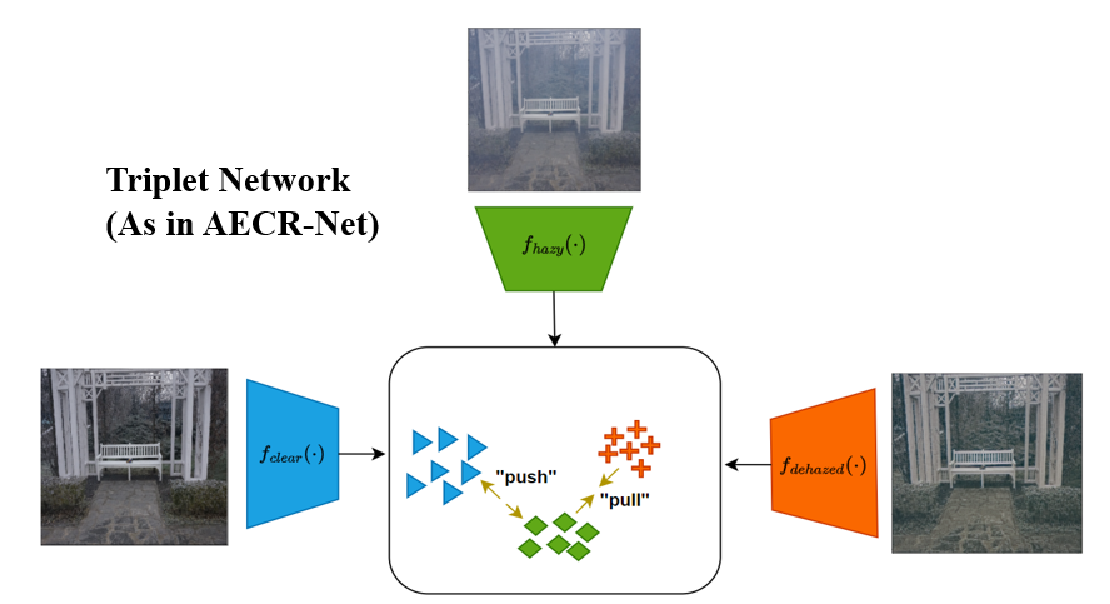}
    \caption{Illustration of the triplet network-based contrastive paradigm (as laid out by the AECR-Net) on the anchor (dehazed), positive (GT) and negative (hazy) image sample.}
    \label{triplet}
\end{figure}

\begin{figure}[hbt!]
    \centering
    \includegraphics[scale=0.80]{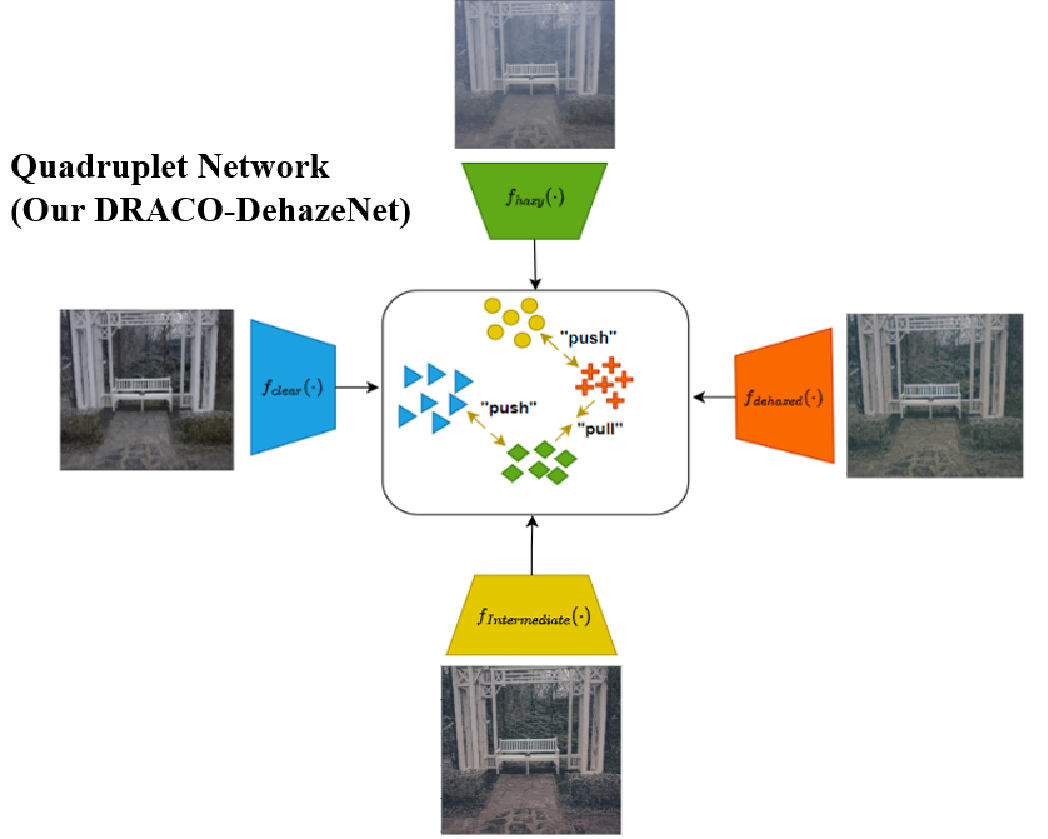}
    \caption{Illustration of our quadruplet network-based contrastive paradigm (as laid out by the AECR-Net) on the anchor (dehazed), intermediate dehazed output, positive (GT) and negative (hazy) image sample.}
    \label{quadruplet}
\end{figure}

The intermediate outputs are interpreted along with the anchor, hazy, and clear images, and hence a quadruplet of images is created, necessitating the use of the quadruplet network and consequently the quadruplet loss. Unlike other related works, our proposed paradigm compute the respective distances via L1 (or Mean Absolute Error (MAE)) loss, as past simulations (e.g., \cite{lan2022online}) have demonstrated that such losses enhance the dehazing output more significantly than the commonly utilized L2 (or Mean Sqaured Error (MSE)) loss, as emphasized by \cite{wu2021contrastive} and \cite{zhao2016loss}. The triplet network and our quadruplet network contrastive dehazing paradigm are illustrated in Figure \ref{triplet} and \ref{quadruplet} respectively to distinguish both approaches.


\subsection{Loss Functions}

Two of the loss functions considered are typical in the majority of dehazing algorithms, namely the Mean Absolute Error (MAE) loss $\mathcal{L}_{mae}$ and the structural similarity (ssim) loss $\mathcal{L}_{ssim}$. The MAE (or L1) loss is mathematically defined as 

\begin{equation}
    \mathcal{L}_{mae} = ||J(x) - I(x) ||_{1},
\end{equation}

whereas the ssim loss $\mathcal{L}_{ssim}$ (more specifically defined as the negative ssim loss) is described mathematically as 

\begin{equation}
    \mathcal{L}_{ssim} = -SSIM(J(x), I(x)), \\
\end{equation}

where $SSIM(J(x), I(x))$ can be explicitly written as 

\begin{equation}
    SSIM = \frac{(C_{1} + 2\mu_{I}\mu_{J})(C_{2} + 2\sigma_{IJ})}{(C_{1} + \mu_{J}^{2} + \mu_{I}^{2})(C_{2} + \sigma_{J}^{2} + \sigma_{I}^{2})}. \\
\end{equation}

In the above, $\mu_{J}$ and $\mu_{I}$ represent the mean values of $J(x)$ and $I(x)$ respectively, $\sigma_{J}$ and $\sigma_{I}$ are the standard deviations of $J(x)$ and $I(x)$ respectively; $\sigma_{IJ}$ is the covariance and $C_{1}$ and $C_{2}$ are constants set to avoid instability, as highlighted by Bergmann et al. \cite{bergmann2018improving}. 

Finally, we propose a quadruplet contrastive loss in our network. It is described by 

\begin{equation}
    \mathcal{L}_{quadruplet} = 
    \mathcal{B} \sum_{i=1}^{N} \Omega \left(\frac{L1_{(J(x),GT(x))}}{L1_{(J(x), I(x))}+ L1_{(J'(x), I(x))} + L1_{(J'(x), J(x))}}\right),
\end{equation}

where $\mathcal{B}$ and $\Omega$ denote the two hyperparameters for balancing the weights of the respective losses during training and $G_{i}(\cdot), i = 1,2,...N$ represents the \emph{i}$\textsuperscript{th}$ hidden features from the respective feature extraction architecture, $GT(x)$ denotes the ground-truth (clear) image, and $J'(x)$ denotes the intermediate dehazed outputs from the DDIRB. For the respective L1 distances: $L1_{(J(x),GT(x))} = ||G_{i}(J(x)) - G_{i}(GT(x))||_{1}$, $L1_{(J(x), I(x))} = ||G_{i}(J(x)) - G_{i}(I(x))||_{1}$, $L1_{(J'(x), I(x))} = ||G_{i}(J'(x)) - G_{i}(I(x))||_{1}$ and $ L1_{(J'(x), J(x))} = ||G_{i}(J(x)) - G_{i}(J'(x))||_{1}$. 

As the latter three terms are in the denominator, they act as a ``pushing'' term that distinguishes between the intermediate output features from the DDIRB and the original hazy feature, as well as between the intermediate output features and the anchor. For comparison, the differences between our quadruplet loss and the triplet loss in AECR-Net is the presence of the additional terms in the denominators, specifically $L1_{(J'(x), I(x))}$ and $L1_{(J'(x), J(x))}$. The triplet loss is mathematically reiterated here as

\begin{equation}
    \mathcal{L}_{triplet} = 
    \mathcal{B} \sum_{i=1}^{N} \Omega \left(\frac{L1_{(J(x),GT(x))}}{L1_{(J(x), I(x))}}\right) = \mathcal{B} \sum_{i=1}^{N} \Omega \left(\frac{||G_{i}(J(x)) - G_{i}(GT(x))||_{1}}{||G_{i}(J(x)) - G_{i}(I(x))||_{1}}\right).
\end{equation}

The additional two terms serve as additional regularization constraint that further refines the clustering capability of the model on the respective embedded features, thereby boosting the dehazing performances.

Overall, the loss function of our DRACO-DehazeNet ($\mathcal{L}_{Draco}$) is the linear summation of $\mathcal{L}_{ssim}$, $\mathcal{L}_{mae}$ and $\mathcal{L}_{quadruplet}$,

\begin{equation}
    \mathcal{L}_{Draco} = \lambda_{1} \cdot \mathcal{L}_{mae} + \lambda_{2} \cdot \mathcal{L}_{ssim} + \lambda_{3} \cdot \mathcal{L}_{quadruplet},
\end{equation}

where $\lambda_{1}$, $\lambda_{2}$ and $\lambda_{3}$ are the weight factors for each loss function, set as 1.0, 1.0 and 0.1, respectively. The two hyperparameters $\mathcal{B}$ and $\Omega$ are set to 0.1 and 0.03125 respectively, which are the same as that of the AECR-Net, since our proposed quadruplet loss function is an extension of the latter's contrastive regularization approach.

\section{Experimental Setup and Settings}

We experimented with and evaluated our proposed approaches relative to the selected SOTAs on 1 synthetic haze and 3 real-world haze dataset, all of which are part of the datasets mentioned in the introductory section. For synthetic haze, we selected the Indoor Training Set (ITS), Outdoor Training Set (OTS) and Synthetic Objective Testing Set (SOTS) of the RESIDE. Following the same train-test paradigm as that in \cite{hong2020distilling}, \cite{qin2020ffa}, \cite{liu2019griddehazenet} and \cite{dong2020multi}, we selected ITS and SOTS as our training and testing datasets, respectively. For real haze, we selected the O-HAZE, NH-HAZE and DENSE-HAZE. 

\subsection{RESIDE}

The RESIDE dataset is one of the most widely used benchmarks for single-image dehazing. It contains a large-scale training and testing image set comprising of indoor and outdoor scenery, sub-categorized into different respective subsets. Specifically, the subsets include ITS, OTS, and SOTS, whose haze is generated synthetically using the ASM, Real World task-driven Testing Set (RTTS), and Hybrid Subjective Testing Set (HSTS), which contain real-world haze scenery. The ITS, OTS and SOTS consists of 13,990, 72,135 and 500 images respectively, while the HSTS and RTTS contained 20 and 4,322 images respectively. Although our main goal is to evaluate our proposed approaches for small real hazy-clear data sample scenarios, the synthetic component of RESIDE is still selected as the benchmark in our study to ensure our algorithm's generalization to the typical synthetic haze scenario with image abundance.  

\subsection{O-HAZE and NH-HAZE}

The O-HAZE dataset comprises of 55 different haze scenes captured outdoors using a real haze generation machine, along with their corresponding counterparts. Following previous related studies, we selected 45 O-HAZE pairs for training, 5 pairs for validation, and the remaining 5 pairs for testing. The NH-HAZE dataset (more specifically NH-HAZE 2020) comprises of 45 different non-homogeneous haze scenes captured outdoors using (again) a real-haze generation machine. Again following previous works, we selected 35 NH-HAZE pairs for training, 5 pairs for validation, and the remaining 5 pairs for testing.

\subsection{DENSE-HAZE}

The DENSE-HAZE dataset comprises of 45 different dense haze scenes from outdoor environments using a real haze generation machine. Unlike O-HAZE and NH-HAZE, this is a relatively difficult dataset to achieve SOTA effective dehazing performance, as pointed out by the original authors. Similar to NH-HAZE, we selected 35 NH-HAZE pairs for training, 5 pairs for validation, and the remaining 5 pairs for testing.



\subsection{Settings}

All experiments were conducted using the Tesla A100 Graphical Processing Units (GPU) from Google Colab, with Tensorflow and Keras as the underlying libraries. The Adam optimizer was utilized for the stochastic gradient descent, and the learning rate of our algorithm was set to 0.001. The epoch selected was 200, with a batch size of 16 (also the same as that of the AECR-Net work) for all the datasets. 


\section{Results and Discussions}
We conducted thorough benchmark tests to verify how well our DRACO-DehazeNet architecture performs compared to the latest dehazing methods. Using well-known image clarity metrics and analyzing computational costs, we assessed its performance on various synthetic and real-world haze datasets. These datasets differed in terms of density, consistency, and size of haze.

Our tests did not only look at the overall model performance; they also examined the impact of each specific component in DRACO-DehazeNet. We performed this procedure through detailed ablation studies. In addition, we visually inspected the respetive output to assess how well the model preserved the details and colors. Through these extensive evaluations, we show that DRACO-DehazeNet efficiently restores image quality, even under limited training data and in challenging haze conditions, and outperforms current methods.

\subsection{On Dehazing Effectiveness}
For all of the benchmarked datasets, the methods selected include DCP, AOD-Net, DehazeNet, FFA-Net, GridDehazeNet (GDN), GFN \cite{ren2018gated}, MSBDN \cite{dong2020multi}, KDDN \cite{hong2020distilling}, FDU \cite{dong2020physics}, Dehamer, SCANet, C$^{2}$PNet, WaveletFormerNet (WFN), and the AECR-Net (which contains the contrastive learning procedure), and served as the SOTA algorithms for comparison with our approach.

\begin{table*} 
\centering
\caption{Quantitative comparisons (PSNR and SSIM) with the SOTA dehazing approaches on RESIDE SOTS (synthetic) and NH-HAZE and DENSE-HAZE (real world hazy). The bolded values depicts the best obtained values, and the underlined values represents the second-best value obtained for each dataset.}
\begin{tabular}{p{4.2cm}p{1.2cm}p{1.2cm}p{1.2cm}p{1.2cm}p{1.2cm}p{1.2cm}}
\hline
& \multicolumn{6}{c}{\textbf{Datasets}} \\ \hline
& \multicolumn{2}{|c|}{\textbf{SOTS}} & \multicolumn{2}{|c|}{\textbf{NH-HAZE}} & \multicolumn{2}{|c|}{\textbf{DENSE-HAZE}} \\ 
\hline
\textbf{Methods} & \textbf{PSNR$\uparrow$} & \textbf{SSIM$\uparrow$} & \textbf{PSNR$\uparrow$} & \textbf{SSIM$\uparrow$} & \textbf{PSNR$\uparrow$} & \textbf{SSIM$\uparrow$} \\
\hline
DCP (\cite{he2010single}, 2010) & 16.62 & 0.8180 & 12.72 & 0.5190 & 11.01 & 0.4170 \\
DehazeNet (\cite{cai2016dehazenet}, 2016) & 19.82 & 0.8210 & 12.72 & 0.5190 & 11.84 & 0.4300  \\
AOD-Net (\cite{li2017aod}, 2017) & 20.51 & 0.8160 & 16.69 & 0.6060 & 12.82 & 0.4680 \\
FFA-Net (\cite{qin2020ffa}, 2020) & 36.39 & 0.9890 & 18.13 & 0.6470 & 14.06 & 0.4520 \\
GDN (\cite{liu2019griddehazenet},2019) & 32.16 & 0.9840 & 16.12 & 0.5950 & 13.60 & 0.4140 \\
GFN (\cite{ren2018gated}, 2018) & 22.30 & 0.8800 & 15.90 & 0.5740 & 12.52 & 0.4240\\
MSBDN (\cite{dong2020multi}, 2020) & 33.67 & 0.9850 & 17.54 & 0.5810 & 14.18 & 0.4110 \\
KDDN (\cite{hong2020distilling}, 2020) & 34.72 & 0.9845 & 17.39 & 0.5897 & 14.28 & 0.4074 \\
FDU (\cite{dong2020physics}, 2020) & 32.68 & 0.9760 & - & - & - & - \\
Dehamer (\cite{guo2022image}, 2022) & 36.63 & 0.9880 & 20.66 & 0.6840 & 16.62 & 0.5600 \\
AECR-Net (\cite{wu2021contrastive}, 2021) & 37.17 & \underline{0.9900} & 19.88 & 0.7170 & 15.91 & 0.4960 \\
SCANet (\cite{guo2023scanet},2023) & - & - & 19.52 & 0.6488 & - & -\\
C$^{2}$PNet (\cite{zheng2023curricular}, 2023) & 36.68 & \underline{0.9900} & 21.32 & \textbf{0.8250} & \underline{16.88} & 0.5730\\
WFN (\cite{zhang2024waveletformernet}, 2024) & 35.96 & 0.9870 & \underline{21.68} & \underline{0.8220} & \textbf{16.95} & \underline{0.5930}\\
\textbf{DRACO-DehazeNet} & \textbf{38.08} & \textbf{0.9906} & \textbf{21.82} & 0.7582 & 14.25 & \textbf{0.6028} \\
 \hline
\end{tabular}
\label{tab:Compare_Tables3}
\end{table*}

\begin{table*} 
\centering
\caption{Quantitative comparisons (PSNR and SSIM) with the SOTA dehazing approaches on O-HAZE (real world hazy). The bolded values depicts the best obtained values, and the underlined values represents the second-best value obtained for each dataset.}
\begin{tabular}{p{4.2cm}p{3.0cm}p{3.0cm}}
\hline
\multicolumn{3}{c}{\textbf{O-HAZE}} \\ 
\hline
\textbf{Methods} & \textbf{PSNR($\uparrow$)} & \textbf{SSIM($\uparrow$)} \\
\hline
DCP (\cite{he2010single},2010) & 14.68 & 0.5200 \\
DehazeNet (\cite{cai2016dehazenet},2016) & 14.65 & 0.5100 \\
AOD-Net (\cite{li2017aod},2017) & 15.07 & 0.5400 \\
FFA-Net (\cite{qin2020ffa},2020) & 17.52 & 0.6140 \\
GFN (\cite{ren2018gated}, 2018) & 14.31 & 0.5310 \\
GDN (\cite{liu2019griddehazenet},2019) & 16.53 & 0.5500\\
MSBDN (\cite{dong2020multi},2020) & \textbf{24.36} & \underline{0.7500} \\
KDDN (\cite{hong2020distilling}, 2020) & 20.62 & 0.7082\\
FDU (\cite{dong2020physics}, 2020) & 20.55 & 0.7157\\
SCANet (\cite{guo2023scanet},2023) & 21.15 & 0.7189 \\
Dehamer (\cite{guo2022image},2022) & 19.47 & 0.7020 \\
AECR-Net (\cite{wu2021contrastive}, 2021) & 19.06 & 0.6370 \\
C$^{2}$PNet (\cite{zheng2023curricular}, 2023) & 20.83 & 0.6920\\
WFN (\cite{zhang2024waveletformernet}, 2024) & 21.32 & 0.7200 \\
\textbf{DRACO-DehazeNet} & \underline{22.94} & \textbf{0.9000} \\
 \hline
\end{tabular}
\label{tab:Compare_Tables5}
\end{table*}

\begin{figure}[hbt!]
    \centering
    \includegraphics[scale=0.66]{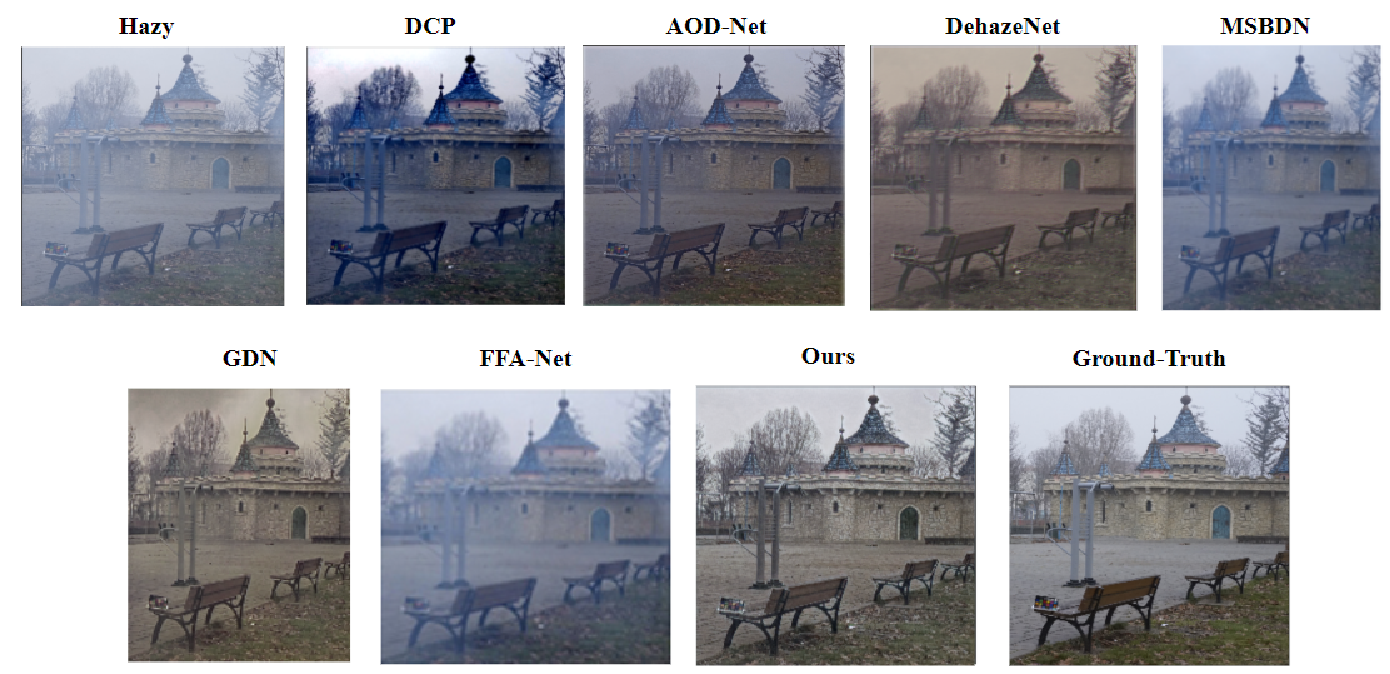}
    \caption{Comparative visual illustration of some of the various dehazed output on a selected O-HAZE image for the SOTAs, including our approach. All the image outputs from the SOTAs are adapted from \cite{jin2022structure}, with permission from Springer Nature.}
    \label{O_haze_compare}
\end{figure}

\begin{figure}[hbt!]
    \centering
    \includegraphics[scale=0.56]{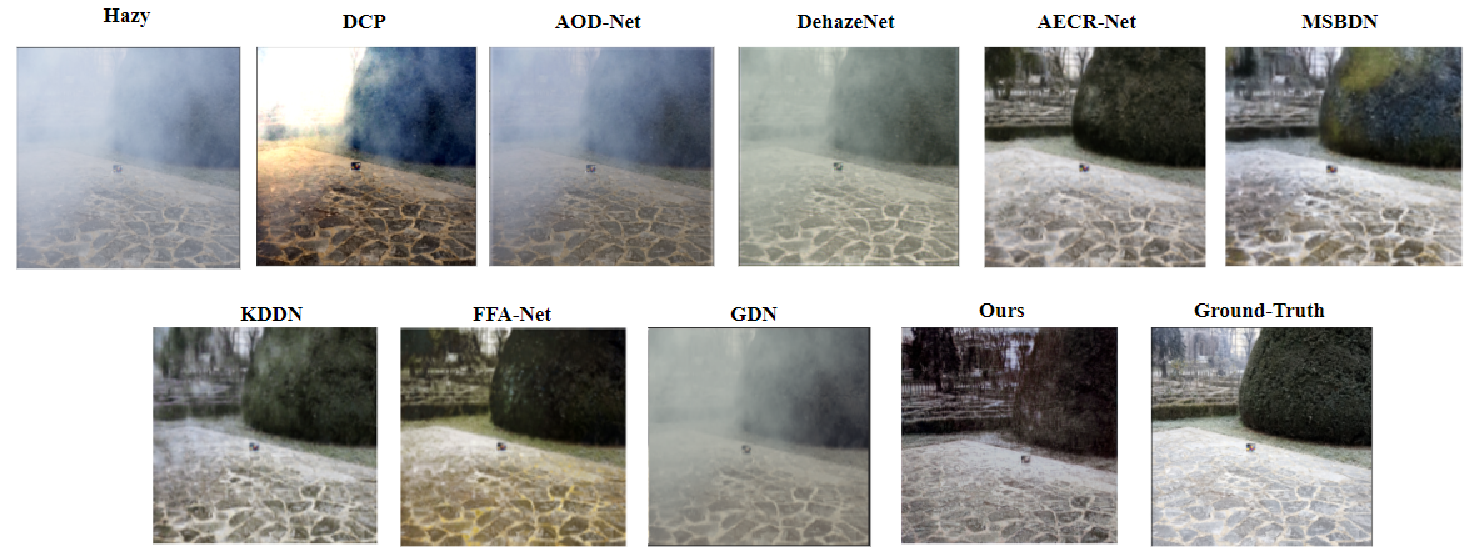}
    \caption{Comparative visual illustration of some of the various dehazed output on a selected NH-HAZE image for the SOTAs, including our approach. All the image outputs from the SOTAs are adapted from \cite{wu2021contrastive}, with permission from IEEE.}
    \label{NH_haze_compare}
\end{figure}

\begin{figure}[hbt!]
    \centering
    \includegraphics[scale=0.56]{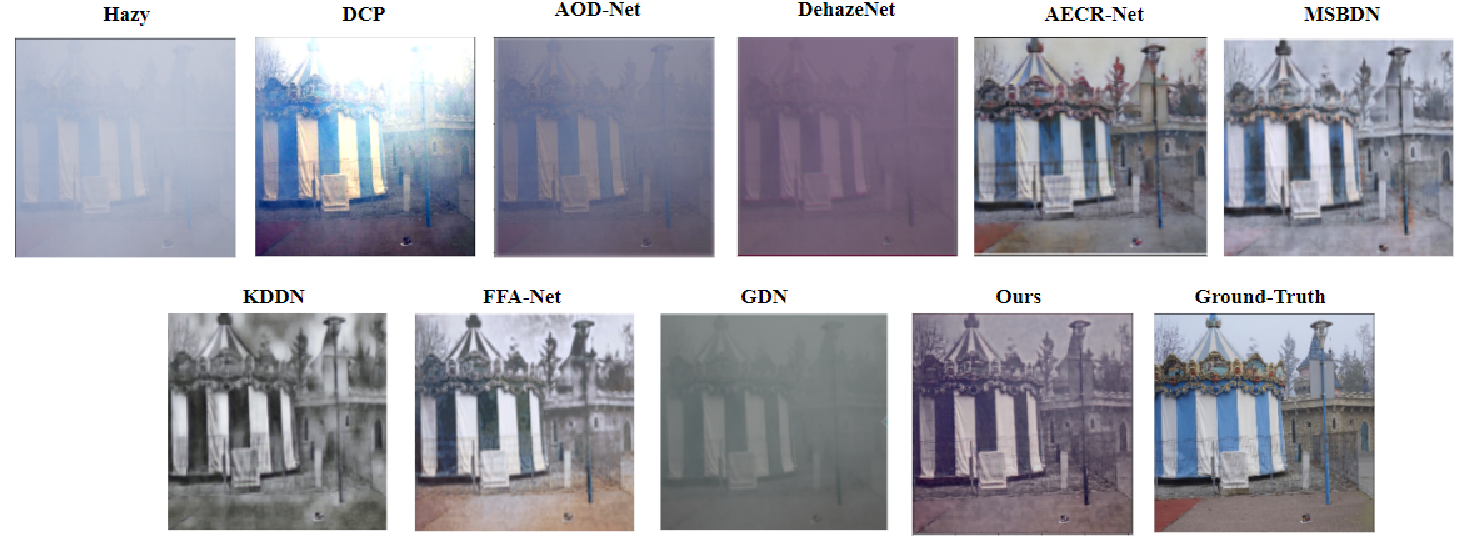}
    \caption{Comparative visual illustration of some of the various dehazed output on a selected DENSE-HAZE image for the SOTAs, including our approach. All the image outputs from the SOTAs are adapted from \cite{wu2021contrastive}, with permission from IEEE.}
    \label{DENSE_haze_compare}
\end{figure}

\begin{figure}[hbt!]
    \centering
    \includegraphics[scale=0.56]{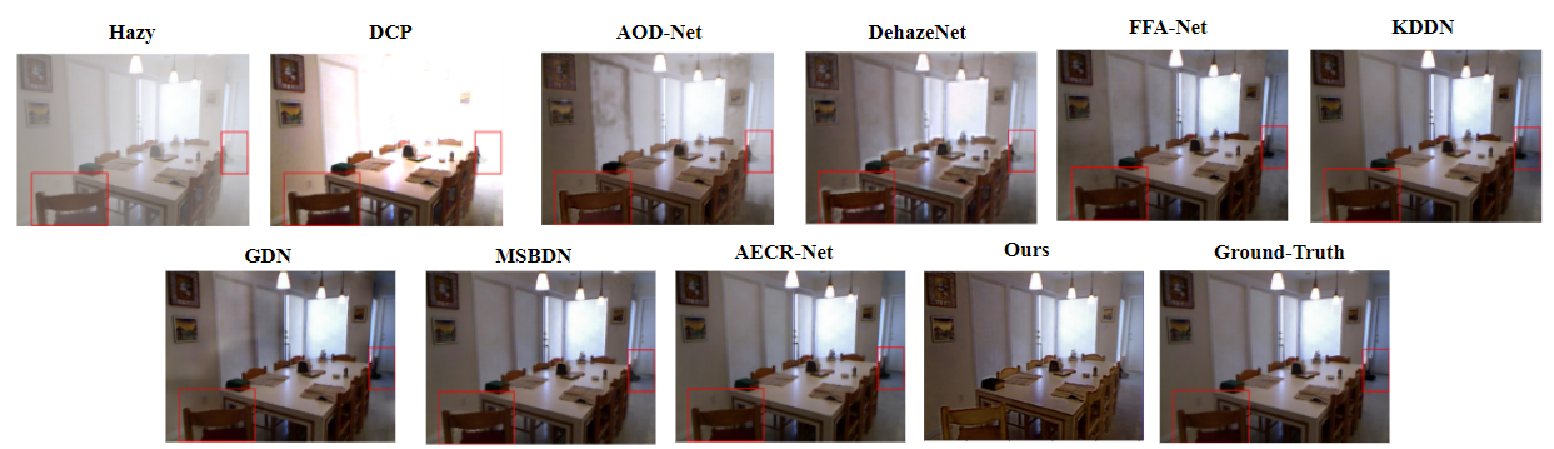}
    \caption{Comparative visual illustration of the various dehazed output on a selected RESIDE (Indoor) image for the SOTAs, including our approach. All the image outputs from the SOTAs are adapted from \cite{wu2021contrastive}, with permission from IEEE.}
    \label{reside_compare}
\end{figure}

Table 3 shows the PSNR and SSIM scores for the selected SOTAs as well as our DRACO-DehazeNet for the SOTS, NH-HAZE, and DENSE-HAZE, while Table 4 shows the corresponding scores obtained using the O-HAZE. The year in which each method was proposed and published is also displayed on the side. We can observed from Table 3 that our method surpassed the SOTAs for the RESIDE in both PSNR and SSIM, for the NH-HAZE in PSNR (with WFN and C$^{2}$PNet outperforming our method in terms of the SSIM), and for the DENSE-HAZE dataset in SSIM, with the AECR-Net, Dehamer, and KDDN outperforming our method in terms of PSNR for the DENSE-HAZE. We can also see from Table 4 (O-HAZE analysis) that our approach surpasses all the selected SOTAs in terms of both SSIM, but for the PSNR, MSBDN surpassed ours. Overall, for all the datasets evaluated, this implies that our approach yielded a more visually pleasing dehazed output, which elucidates the role of our contrastive paradigm in refining the dehazing outputs visually to a new level. This is justified by Figure \ref{O_haze_compare}, which depicts the respective dehazed outputs for each SOTA for a selected O-HAZE test image. As an example comparison, the MSBDN dehazed output still contained some amount of haze in the background, whereas the GDN dehazed output yielded significant color distortions and contrasts, thus explaining their relatively lower SSIM values. Our approach yielded the least contrast and color deviation relative to the ground-truth reference, compared to the other methods. A similar observation was reported for NH-HAZE, whereby the SSIM obtained for our approach is the most optimal. This is also illustrated in Figure \ref{NH_haze_compare}, where our method once again yielded the output that deviates the least from the ground-truth in terms of contrasts and color distortions. This can be seen by observing that for the other methods, the floor tile colors are over-saturated and over-contrasted relative to the ground-truth image reference. These observations illustrate the capability of incorporating the detail recovery network to enhance the visual appearance of the dehazed output, Such a network has also been shown to be effective for non-homogeneous real haze apart from the widely tested homogeneous real haze, the first of its kind in such a scenario. For DENSE-HAZE, our method, like the rest of the SOTAs, failed to achieve satisfactory dehazing results, despite attaining the highest SSIM values. As agreed with \cite{ancuti2019dense}, the original creator of the dataset, as well as \cite{wu2021contrastive} and \cite{jin2022structure}, the thicker haze scenario leads to severe degradation of information and hence is significantly more difficult to remove than the haze in NH-HAZE and O-HAZE. The various dehazed outputs obtained from the DENSE-HAZE and RESIDE images are shown in Figures \ref{DENSE_haze_compare} and \ref{reside_compare}, respectively. 

\subsection{Ablation Studies}

\begin{figure}[hbt!]
    \centering
    \includegraphics[scale=0.65]{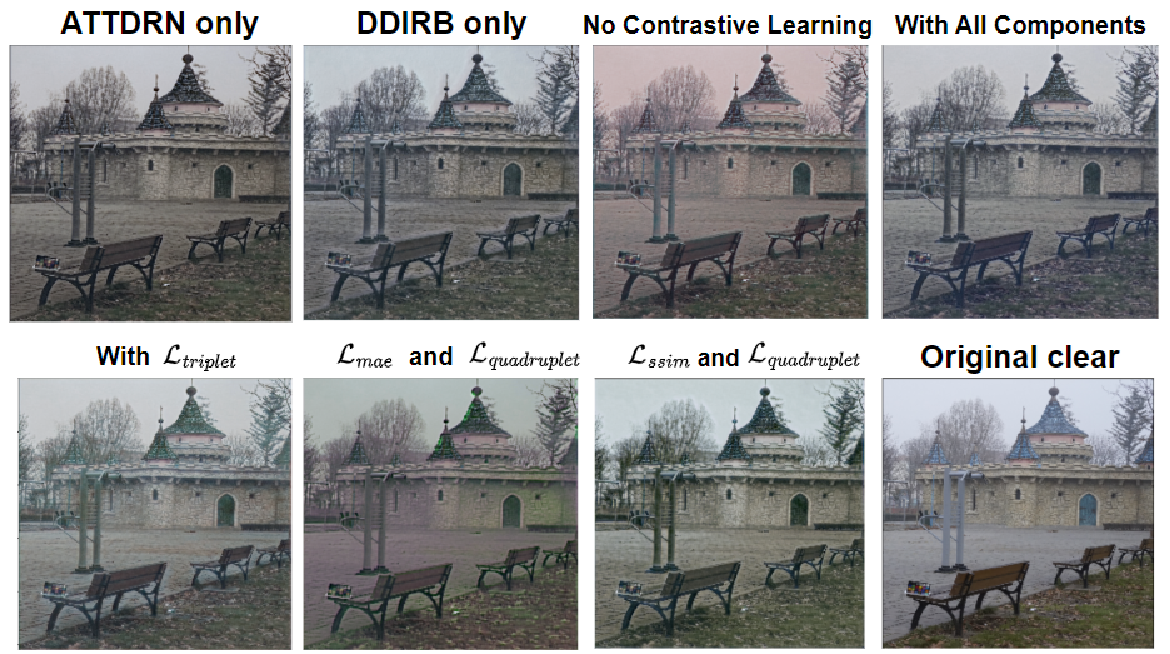}
    \caption{Comparative illustration of the respective algorithmic component ablation (top) and loss function ablation (bottom) for our approach as outlined in Table 6 and 7 for an O-HAZE image. The original clear image is also illustrated for reference.}
    \label{Ablation_O_Haze}
\end{figure}

To demonstrate the effectiveness of the various novel components of our DRACO-DehazeNet, we conducted an ablation study to analyze different elements, including the inverted residual block in DDIRB, the ATTDRN, the contrastive learning component, and the various loss functions. More specifically, we split our ablation analysis into two parts: The algorithmic component ablation, and the loss function ablation. For the algorithmic component ablation, we utilized the following configurations:  DRACO-DehazeNet with DDIRB only, DRACO-DehazeNet with ATTDRN only, and DRACO-DehazeNet without contrastive learning. For the loss function ablation, we make use of the original DRACO-DehazeNet configuration $\mathcal{L}_{Draco} = \lambda_{1} \cdot \mathcal{L}_{mae} + \lambda_{2} \cdot \mathcal{L}_{ssim} + \lambda_{3} \cdot \mathcal{L}_{quadruplet}$ and experimented on training it via the $\lambda_{1} \cdot \mathcal{L}_{mae} + \lambda_{3} \cdot \mathcal{L}_{quadruplet}$ and $\lambda_{2} \cdot \mathcal{L}_{ssim} + \lambda_{3} \cdot \mathcal{L}_{quadruplet}$ configuration. To compare the performance of our proposed quadruplet loss relative to the triplet loss $\mathcal{L}_{triplet}$, the configuration $\lambda_{1} \cdot \mathcal{L}_{mae} + \lambda_{2} \cdot \mathcal{L}_{ssim}+ \lambda_{3} \cdot \mathcal{L}_{triplet}$ was also evaluated, where $\lambda_{3}$ remained at 0.1. The illustrated numerical results are tabulated in Tables 5 and 6 using O-HAZE, along with Figure \ref{Ablation_O_Haze} depicting the individual dehazed output for the respective ablation configuration for a particular image. 

For the algorithmic ablation analysis, we can immediately notice is that removing either DDIRB or ATTDRN module in our DRACO-DehazeNet would reduce both the PSNR and SSIM metrics values, with the ATTDRN component-only configuration yielding lesser metric values than with the DDIRB component only. This is justified in Figure 8 in which the ATTDRN-only dehazed output has more color distortion and contrast relative to the DDIRB-only output. This means that the ATTDRN by itself would not yield a significantly pleasing dehazing output, and thus emphasized the role of its incorporation into the DDIRB quantitatively to enhance the dehazing performance via detail recovery. It should be noted that, without utilizing contrastive learning, the PSNR and SSIM values decreases by 6.93$\%$ and 0.28$\%$ respectively, compared to the original configuration. This highlights the benefit of incorporating contrastive network to guide the extracted dehazed features to cluster near the ground-truth clear features and away from the hazy features, further complementing the framework set by \cite{park2020contrastive} on utilizing contrastive learning to enhance unpaired image-to-image translation quality, as also agreed by \cite{wu2021contrastive}.

\begin{table*} 
\centering
\caption{Comparison of the role of algorithmic components used in the training of DRACO-DehazeNet on O-HAZE. The bolded values depicts the best obtained values.}
\begin{tabular}{p{5cm}p{3.0cm}p{3.0cm}}
\hline
\multicolumn{3}{c}{\textbf{O-HAZE Ablation (Algorithmic Components)}} \\ 
\hline
\textbf{Configuration} & \textbf{PSNR($\uparrow$)} & \textbf{SSIM($\uparrow$)} \\
\hline
 All 3 Components &  \textbf{22.94} & \textbf{0.9000}  \\
 DDIRB only & 21.18 & 0.8890 \\
 ATTDRN only & 20.34 & 0.8791\\
 No Contrastive Learning & 21.35 & 0.8974 \\
 \hline
\end{tabular}
\label{tab:Ablation1}
\end{table*}

\begin{table*} 
\centering
\caption{Comparison of the role of loss functions used in the training of DRACO-DehazeNet on O-HAZE. The configuration used is the original DRACO-DehazeNet. The bolded values depicts the best obtained values.}
\begin{tabular}{p{7cm}p{3.0cm}p{3.0cm}}
\hline
\multicolumn{3}{c}{\textbf{O-HAZE Ablation (Loss Functions)}} \\ 
\hline
\textbf{Configuration} & \textbf{PSNR($\uparrow$)} & \textbf{SSIM($\uparrow$)} \\
\hline
 $\lambda_{1} \cdot \mathcal{L}_{mae} + \lambda_{2} \cdot \mathcal{L}_{ssim}+ \lambda_{3} \cdot \mathcal{L}_{quadruplet}$ & \textbf{22.94} & \textbf{0.9000}  \\
 $\lambda_{1} \cdot \mathcal{L}_{mae} +  \lambda_{3} \cdot \mathcal{L}_{quadruplet}$ & 19.20 & 0.8365 \\
 $\lambda_{2} \cdot \mathcal{L}_{ssim} +  \lambda_{3} \cdot \mathcal{L}_{quadruplet}$ & 22.63 & 0.8912 \\
 $\lambda_{1} \cdot \mathcal{L}_{mae} + \lambda_{2} \cdot \mathcal{L}_{ssim} + \lambda_{3} \cdot \mathcal{L}_{triplet}$ & 22.54& 0.8917 \\
 \hline
\end{tabular}
\label{tab:Ablation2}
\end{table*}

For the loss function ablation analysis, we can see that incorporating $\mathcal{L}_{ssim}$ is more critical than incorporating $\mathcal{L}_{mae}$, as the former allowed a more visually pleasing dehazed output of a lighter contrast compared to the latter, as shown in Figure \ref{Ablation_O_Haze}. Nevertheless, it is by using all three loss functions simultaneously for $\mathcal{L}_{Draco}$ that we can achieve a higher PSNR and SSIM values. Finally, we can see that utilizing $\mathcal{L}_{quadruplet}$ indeed agreed with our overall hypothesis of providing a better dehazing refinement via a quadruplet network than that of the triplet network using $\mathcal{L}_{triplet}$, albeit not by a significant amount of 0.92$\%$. Such a close comparative dehazing output can be seen in Figure \ref{Ablation_O_Haze} where the visual output for the triplet network (leftmost of the second row of the diagram) has only a slight color contrast difference to that of our approach, yet the output for the latter resembles that of the original clear image, justifying the higher PSNR and SSIM obtained.

\subsection{On Dehazing Efficiency}


We have also evaluated the efficiency of our algorithm in terms of the FLOPs. Although our proposed design may not be as efficient as the other simpler CNN models such as DehazeNet and AOD-Net, it has already demonstrated superior dehazing performances visually. Hence, our primary goal now is to ensure that it is at least computationally less demanding than the closely related DPE-Net structure, which comprises of the DDRB and ERPAB networks. Table 7 illustrates the FLOPs for our approach relative to DPE-Net on the O-HAZE dataset. We can see that for our approach, the FLOPs obtained are lower than that of the DPE-Net by 61.8 $\%$. Specifically, via an analogous ablation analysis approach using the configurations highlighted in Table 8, we can see that utilizing only the DDIRB component in our DRACO-DehazeNet results in significantly lower FLOPs requirement. However, using the ATTDRN only in our DRACO-DehazeNet required more FLOPs than using the DDIRB, as our ATTDRN architecture consisted of concatenating the dilated convolution from three different dilation rates before passing them to both the channel and spatial attention. Nevertheless, the overall FLOPs required for our network are lesser than those of DPE-Net, and this empirically demonstrated that the inverted residual block helps to significantly lower the computational cost in our approach. Our analysis also opens up the possibility of making such contrastive-based detail recovery dehazing architecture more efficient, and we foresee further developments in the design of both effective and efficient dehazing algorithm, particularly as their applications have been increasingly bought over to mobile platforms.

\begin{table*} 
\centering
\caption{Comparison of overall computational efficiency in terms of FLOPs on the O-HAZE for our DRACO-DehazeNet versus the related DPE-Net. The bolded values depicts the best obtained values.}
\begin{tabular}{p{5cm}p{3.0cm}p{3.0cm}}
\hline
\multicolumn{3}{c}{\textbf{O-HAZE Algorithmic Efficiency (Overall)}} \\ 
\hline
\textbf{Configuration} & \textbf{FLOPs($\downarrow$)} \\
\hline
 DRACO-DehazeNet  & \textbf{32.4G} \\
 DPE-Net & 84.9G \\
 \hline
\end{tabular}
\label{tab:Efficiency1}
\end{table*}

\begin{table*} 
\centering
\caption{Comparison of the role of algorithmic components used in our DRACO-DehazeNet on the computational efficiency in terms of FLOPs on the O-HAZE. The bolded values depicts the optimal values.}
\begin{tabular}{p{5cm}p{3.0cm}p{3.0cm}}
\hline
\multicolumn{3}{c}{\textbf{O-HAZE Algorithmic Efficiency (Components)}} \\ 
\hline
\textbf{Configuration} & \textbf{FLOPs($\downarrow$)} \\
\hline
 DDIRB only & \textbf{1.2G} \\
 ATTDRN only & 29.7G  \\
 No Contrastive Learning & 30.8G\\
 All 3 components & 32.4G \\
 \hline
\end{tabular}
\label{tab:Efficiency2}
\end{table*}

Finally, Table 9 compares our approach to that of the selected SOTA dehazing models in terms of the efficiency and dehazing metrics of the NH-HAZE. We can see that our method yielded lower FLOPs than SCANet, Dehamer, AECR-Net, FFA-Net, and MSBDN, but still yielded higher FLOPs than AOD-Net and GridDehazeNet. However, the latter two approaches yielded significantly lower dehazing metrics than the others, for which the PSNR and SSIM usually lie beyond 17.00 and 0.6800 respectively. Our approach is the only method in the table that utilizes training parameter below 1M and FLOPs below 40G while retaining inside the aforementioned PSNR and SSIM bounds. Our proposed method also retained the highest metric values throughout this comparative study. The Dehamer, which is the only ViT-based approach in Table 9, required the highest FLOPs value of 870G, yet yielded metric values that are smaller than the AECR-Net and our approach, which only required FLOPs value of 43.0G and 32.4G respectively. This, in a way, empirically justified the superiority of contrastive-based dehazing models as compared to ViT-based dehazing (and a majority of CNN-based) models in consuming lesser computational resources while achieving higher dehazing performances.  

\begin{table}
\centering
\caption{List of some selected dehazing approach with their number of training parameters and FLOPs required, as well as the PSNR and SSIM values, evaluated on the NH-HAZE. The bolded value depicts the optimal value, while the underlined value depicts the second-best value.}
\begin{tabular}{p{3cm}p{2.5cm}p{2cm}p{2cm}p{2cm}}
\hline
\textbf{Models} & \textbf{Params($\downarrow$)} & \textbf{FLOPs($\downarrow$)} & \textbf{PSNR($\uparrow$)} & \textbf{SSIM($\uparrow$)}
\\
\hline
AOD-Net & \textbf{0.002M} & \textbf{0.1G} & 15.40 & 0.5693 \\
GridDehazeNet & 0.96M & \underline{21.5G} & 13.80 & 0.5370 \\
FFA-Net & 4.68M & 288.1G & 19.87 & 0.6915 \\
Dehamer & 29.44M & 870G & \underline{20.66} & 0.6844 \\
MSBDN & 31.35M & 41.5G & 19.23 & 0.7056 \\
AECR-Net & 2.61M & 43.0G & 19.88 & \underline{0.7173}\\
SCANet & 2.39M & 258.6G & 19.52 & 0.6488\\
\textbf{Ours} & \underline{0.30M} & 32.4G & \textbf{21.82} & \textbf{0.7582}\\
\hline
\end{tabular}
\label{tab:Efficiency_comparison}
\end{table}

\section{Conclusions}

We have proposed and implemented DRACO-DehazeNet, a novel dehazing network that emphasizes efficiency and effectiveness by incorporating a dense dilated inverted residual block module, detail recovery via the attention-based detail recovery module, and quadruplet network-based contrastive learning to guide the dehazing procedure. Our network is the first of its kind to incorporate the inverted residual block in its dehazing architecture design, and when used in tandem with the other aforementioned modules, boosted the dehazing effectiveness relative to the SOTAs to a new level in general, and also suggests that our network enhances the visual appearances of the dehazed output significantly. Our network was subjected to rigorous evaluation procedure involving 3 real-world haze dataset, some with small data sizes to assess the role of quadruplet contrastive dehazing in handling small data samples, and 1 synthetic haze dataset. Our method also demonstrated better algorithmic efficiency via lower FLOPs values relative to the related DPE-Net, which we took inspiration from in the design of our dehazing network, as well as the other selected SOTA dehazing architecture.

\section*{CRediT authorship contribution statement}

\textbf{Gao Yu Lee}: Conceptualization, Investigation, Methodology, Software, Writing - original draft, Writing- review $\&$ editing. \textbf{Tanmoy Dam}:  Conceptualization, Investigation, Methodology, Supervision, Writing- review $\&$ editing. \textbf{Md Mefahul Ferdaus}: Supervision, Writing- review $\&$ editing. \textbf{Daniel Puiu Poenar}: Supervision. \textbf{Vu N. Duong}: Supervision.    

\section*{Declaration of Competing Interest}

The authors declared no competing interests or conflicts that would influenced the work reported in this paper.

\section*{Acknowledgements}
This research/project is supported by the Civil Aviation Authority of Singapore and NTU under their collaboration in the Air Traffic Management Research Institute. Any opinions, findings and conclusions or recommendations expressed in this material are those of the author(s) and do not necessarily reflect the views of the Civil Aviation Authority of Singapore.



 \bibliographystyle{elsarticle-num} 
 \bibliography{ref}

\begin{thebibliography}{10}
\expandafter\ifx\csname url\endcsname\relax
  \def\url#1{\texttt{#1}}\fi
\expandafter\ifx\csname urlprefix\endcsname\relax\def\urlprefix{URL }\fi
\expandafter\ifx\csname href\endcsname\relax
  \def\href#1#2{#2} \def\path#1{#1}\fi

\bibitem{silberman2012indoor}
N.~Silberman, D.~Hoiem, P.~Kohli, R.~Fergus, Indoor segmentation and support inference from rgbd images, in: Computer Vision--ECCV 2012: 12th European Conference on Computer Vision, Florence, Italy, October 7-13, 2012, Proceedings, Part V 12, Springer, 2012, pp. 746--760.

\bibitem{li2018benchmarking}
B.~Li, W.~Ren, D.~Fu, D.~Tao, D.~Feng, W.~Zeng, Z.~Wang, Benchmarking single-image dehazing and beyond, IEEE Transactions on Image Processing 28~(1) (2018) 492--505.

\bibitem{koschmieder1925theorie}
H.~Koschmieder, Theorie der horizontalen Sichtweite, Keim \& Nemnich, 1925.

\bibitem{ancuti2018haze}
C.~Ancuti, C.~O. Ancuti, R.~Timofte, C.~De~Vleeschouwer, I-haze: A dehazing benchmark with real hazy and haze-free indoor images, in: Advanced Concepts for Intelligent Vision Systems: 19th International Conference, ACIVS 2018, Poitiers, France, September 24--27, 2018, Proceedings 19, Springer, 2018, pp. 620--631.

\bibitem{ancuti2018ohaze}
C.~O. Ancuti, C.~Ancuti, R.~Timofte, C.~De~Vleeschouwer, O-haze: a dehazing benchmark with real hazy and haze-free outdoor images, in: Proceedings of the IEEE conference on computer vision and pattern recognition workshops, 2018, pp. 754--762.

\bibitem{ancuti2020nh}
C.~O. Ancuti, C.~Ancuti, R.~Timofte, Nh-haze: An image dehazing benchmark with non-homogeneous hazy and haze-free images, in: Proceedings of the IEEE/CVF conference on computer vision and pattern recognition workshops, 2020, pp. 444--445.

\bibitem{ancuti2019dense}
C.~O. Ancuti, C.~Ancuti, M.~Sbert, R.~Timofte, Dense-haze: A benchmark for image dehazing with dense-haze and haze-free images, in: 2019 IEEE international conference on image processing (ICIP), IEEE, 2019, pp. 1014--1018.

\bibitem{song2023vision}
Y.~Song, Z.~He, H.~Qian, X.~Du, Vision transformers for single image dehazing, IEEE Transactions on Image Processing 32 (2023) 1927--1941.

\bibitem{zhang2020drcdn}
S.~Zhang, F.~He, Drcdn: learning deep residual convolutional dehazing networks, The Visual Computer 36~(9) (2020) 1797--1808.

\bibitem{feng2021urnet}
T.~Feng, C.~Wang, X.~Chen, H.~Fan, K.~Zeng, Z.~Li, Urnet: A u-net based residual network for image dehazing, Applied Soft Computing 102 (2021) 106884.

\bibitem{yang2022transformer}
Z.~Yang, X.~Li, J.~Li, Transformer-based progressive residual network for single image dehazing, Frontiers in Neurorobotics 16 (2022) 1084543.

\bibitem{lan2022online}
Y.~Lan, Z.~Cui, Y.~Su, N.~Wang, A.~Li, W.~Zhang, Q.~Li, X.~Zhong, Online knowledge distillation network for single image dehazing, Scientific Reports 12~(1) (2022) 14927.

\bibitem{hong2020distilling}
M.~Hong, Y.~Xie, C.~Li, Y.~Qu, Distilling image dehazing with heterogeneous task imitation, in: Proceedings of the IEEE/CVF Conference on Computer Vision and Pattern Recognition, 2020, pp. 3462--3471.

\bibitem{wang2022multi}
N.~Wang, Z.~Cui, A.~Li, Y.~Su, Y.~Lan, Multi-priors guided dehazing network based on knowledge distillation, in: Chinese Conference on Pattern Recognition and Computer Vision (PRCV), Springer, 2022, pp. 15--26.

\bibitem{feng2019image}
T.~Feng, Z.~Li, C.~Wang, X.~Chen, J.~Wu, Image dehazing network based on dilated convolution feature extraction, in: 2019 12th International Congress on Image and Signal Processing, BioMedical Engineering and Informatics (CISP-BMEI), IEEE, 2019, pp. 1--5.

\bibitem{deivalakshmi2022deep}
S.~Deivalakshmi, J.~Sudaroli~Sandana, Deep dilated convolutional network for single image dehazing, in: International Conference on Computer Vision and Image Processing, Springer, 2022, pp. 281--291.

\bibitem{kuanar2019night}
S.~Kuanar, K.~Rao, D.~Mahapatra, M.~Bilas, Night time haze and glow removal using deep dilated convolutional network, arXiv preprint arXiv:1902.00855 (2019).

\bibitem{howard2017mobilenets}
A.~G. Howard, M.~Zhu, B.~Chen, D.~Kalenichenko, W.~Wang, T.~Weyand, M.~Andreetto, H.~Adam, Mobilenets: Efficient convolutional neural networks for mobile vision applications, arXiv preprint arXiv:1704.04861 (2017).

\bibitem{sandler2018mobilenetv2}
M.~Sandler, A.~Howard, M.~Zhu, A.~Zhmoginov, L.-C. Chen, Mobilenetv2: Inverted residuals and linear bottlenecks, in: Proceedings of the IEEE conference on computer vision and pattern recognition, 2018, pp. 4510--4520.

\bibitem{tan2019efficientnet}
M.~Tan, Q.~Le, Efficientnet: Rethinking model scaling for convolutional neural networks, in: International conference on machine learning, PMLR, 2019, pp. 6105--6114.

\bibitem{li2022single}
Y.~Li, D.~Cheng, D.~Zhang, N.~Wang, X.~Gao, J.~Sun, Single image dehazing with an independent detail-recovery network, Knowledge-Based Systems 254 (2022) 109579.

\bibitem{fang2022detail}
W.~Fang, Z.~Huo, Y.~Qiao, Detail recovery and color enhancement for single image dehazing, in: Proceedings of the 2022 11th International Conference on Networks, Communication and Computing, 2022, pp. 67--75.

\bibitem{gao2022heavy}
T.~Gao, Y.~Wen, J.~Zhang, K.~Zhang, T.~Chen, From heavy rain removal to detail restoration: A faster and better network, arXiv preprint arXiv:2205.03553 (2022).

\bibitem{deng2020detail}
S.~Deng, M.~Wei, J.~Wang, Y.~Feng, L.~Liang, H.~Xie, F.~L. Wang, M.~Wang, Detail-recovery image deraining via context aggregation networks, in: Proceedings of the IEEE/CVF conference on computer vision and pattern recognition, 2020, pp. 14560--14569.

\bibitem{shen2022detail}
Y.~Shen, M.~Wei, S.~Deng, W.~Yang, Y.~Wang, X.-P. Zhang, M.~Wang, J.~Qin, Detail-recovery image deraining via dual sample-augmented contrastive learning, arXiv preprint arXiv:2204.02772 (2022).

\bibitem{zhu2022hdrd}
D.~Zhu, S.~Deng, W.~Wang, G.~Cheng, M.~Wei, F.~L. Wang, H.~Xie, Hdrd-net: High-resolution detail-recovering image deraining network, Multimedia Tools and Applications 81~(29) (2022) 42889--42906.

\bibitem{ahn2022remove}
W.~J. Ahn, T.~K. Kang, H.~D. Choi, M.~T. Lim, Remove and recover: Deep end-to-end two-stage attention network for single-shot heavy rain removal, Neurocomputing 481 (2022) 216--227.

\bibitem{jiang2023two}
R.~Jiang, Y.~Li, C.~Chen, W.~Liu, Two-stage learning framework for single image deraining, IET Image Processing 17~(5) (2023) 1449--1463.

\bibitem{zhao2021complementary}
D.~Zhao, J.~Li, H.~Li, L.~Xu, Complementary feature enhanced network with vision transformer for image dehazing, arXiv preprint arXiv:2109.07100 (2021).

\bibitem{hoffer2015deep}
E.~Hoffer, N.~Ailon, Deep metric learning using triplet network, in: Similarity-Based Pattern Recognition: Third International Workshop, SIMBAD 2015, Copenhagen, Denmark, October 12-14, 2015. Proceedings 3, Springer, 2015, pp. 84--92.

\bibitem{snell2017prototypical}
J.~Snell, K.~Swersky, R.~Zemel, Prototypical networks for few-shot learning, Advances in neural information processing systems 30 (2017).

\bibitem{lee2024dehazing}
G.~Y. Lee, J.~Chen, T.~Dam, M.~M. Ferdaus, D.~P. Poenar, V.~N. Duong, Dehazing remote sensing and uav imagery: A review of deep learning, prior-based, and hybrid approaches, arXiv preprint arXiv:2405.07520 (2024).

\bibitem{wu2021contrastive}
H.~Wu, Y.~Qu, S.~Lin, J.~Zhou, R.~Qiao, Z.~Zhang, Y.~Xie, L.~Ma, Contrastive learning for compact single image dehazing, in: Proceedings of the IEEE/CVF Conference on Computer Vision and Pattern Recognition, 2021, pp. 10551--10560.

\bibitem{zheng2023curricular}
Y.~Zheng, J.~Zhan, S.~He, J.~Dong, Y.~Du, Curricular contrastive regularization for physics-aware single image dehazing, in: Proceedings of the IEEE/CVF conference on computer vision and pattern recognition, 2023, pp. 5785--5794.

\bibitem{cheng2022robust}
D.~Cheng, Y.~Li, D.~Zhang, N.~Wang, X.~Gao, J.~Sun, Robust single image dehazing based on consistent and contrast-assisted reconstruction, arXiv preprint arXiv:2203.15325 (2022).

\bibitem{chen2017beyond}
W.~Chen, X.~Chen, J.~Zhang, K.~Huang, Beyond triplet loss: a deep quadruplet network for person re-identification, in: Proceedings of the IEEE conference on computer vision and pattern recognition, 2017, pp. 403--412.

\bibitem{ali2023lidn}
A.~Ali, A.~Ghosh, S.~S. Chaudhuri, Lidn: a novel light invariant image dehazing network, Engineering Applications of Artificial Intelligence 126 (2023) 106830.

\bibitem{ancuti2020day}
C.~Ancuti, C.~O. Ancuti, C.~De~Vleeschouwer, A.~C. Bovik, Day and night-time dehazing by local airlight estimation, IEEE Transactions on Image Processing 29 (2020) 6264--6275.

\bibitem{ding2022slimyolov4}
P.~Ding, H.~Qian, S.~Chu, Slimyolov4: lightweight object detector based on yolov4, Journal of Real-Time Image Processing 19~(3) (2022) 487--498.

\bibitem{sankararaman2020impact}
K.~A. Sankararaman, S.~De, Z.~Xu, W.~R. Huang, T.~Goldstein, The impact of neural network overparameterization on gradient confusion and stochastic gradient descent, in: International conference on machine learning, PMLR, 2020, pp. 8469--8479.

\bibitem{he2010single}
K.~He, J.~Sun, X.~Tang, Single image haze removal using dark channel prior, IEEE transactions on pattern analysis and machine intelligence 33~(12) (2010) 2341--2353.

\bibitem{fattal2008single}
R.~Fattal, Single image dehazing, ACM transactions on graphics (TOG) 27~(3) (2008) 1--9.

\bibitem{zhu2015fast}
Q.~Zhu, J.~Mai, L.~Shao, A fast single image haze removal algorithm using color attenuation prior, IEEE transactions on image processing 24~(11) (2015) 3522--3533.

\bibitem{wei2021non}
H.~Wei, Q.~Wu, H.~Li, K.~N. Ngan, H.~Li, F.~Meng, L.~Xu, Non-homogeneous haze removal via artificial scene prior and bidimensional graph reasoning, IEEE Transactions on Image Processing 30 (2021) 9136--9149.

\bibitem{cai2016dehazenet}
B.~Cai, X.~Xu, K.~Jia, C.~Qing, D.~Tao, Dehazenet: An end-to-end system for single image haze removal, IEEE transactions on image processing 25~(11) (2016) 5187--5198.

\bibitem{ren2016single}
W.~Ren, S.~Liu, H.~Zhang, J.~Pan, X.~Cao, M.-H. Yang, Single image dehazing via multi-scale convolutional neural networks, in: Computer Vision--ECCV 2016: 14th European Conference, Amsterdam, The Netherlands, October 11-14, 2016, Proceedings, Part II 14, Springer, 2016, pp. 154--169.

\bibitem{li2017aod}
B.~Li, X.~Peng, Z.~Wang, J.~Xu, D.~Feng, Aod-net: All-in-one dehazing network, in: Proceedings of the IEEE international conference on computer vision, 2017, pp. 4770--4778.

\bibitem{qin2020ffa}
X.~Qin, Z.~Wang, Y.~Bai, X.~Xie, H.~Jia, Ffa-net: Feature fusion attention network for single image dehazing, in: Proceedings of the AAAI conference on artificial intelligence, Vol.~34, 2020, pp. 11908--11915.

\bibitem{liu2019griddehazenet}
X.~Liu, Y.~Ma, Z.~Shi, J.~Chen, Griddehazenet: Attention-based multi-scale network for image dehazing, in: Proceedings of the IEEE/CVF international conference on computer vision, 2019, pp. 7314--7323.

\bibitem{zhang2018densely}
H.~Zhang, V.~M. Patel, Densely connected pyramid dehazing network, in: Proceedings of the IEEE conference on computer vision and pattern recognition, 2018, pp. 3194--3203.

\bibitem{wang2022msf}
G.~Wang, X.~Yu, Msf 2 dn: Multi scale feature fusion dehazing network with dense connection, in: Asian Conference on Computer Vision, Springer, 2022, pp. 444--459.

\bibitem{yi2021msnet}
Q.~Yi, A.~Jiang, X.~Deng, C.~Liu, Msnet: A novel end-to-end single image dehazing network with multiple inter-scale dense skip-connections, IET Image Processing 15~(1) (2021) 143--154.

\bibitem{guo2022image}
C.-L. Guo, Q.~Yan, S.~Anwar, R.~Cong, W.~Ren, C.~Li, Image dehazing transformer with transmission-aware 3d position embedding, in: Proceedings of the IEEE/CVF Conference on Computer Vision and Pattern Recognition, 2022, pp. 5812--5820.

\bibitem{wang2018understanding}
P.~Wang, P.~Chen, Y.~Yuan, D.~Liu, Z.~Huang, X.~Hou, G.~Cottrell, Understanding convolution for semantic segmentation, in: 2018 IEEE winter conference on applications of computer vision (WACV), Ieee, 2018, pp. 1451--1460.

\bibitem{luo2023lkd}
P.~Luo, G.~Xiao, X.~Gao, S.~Wu, Lkd-net: Large kernel convolution network for single image dehazing, in: 2023 IEEE International Conference on Multimedia and Expo (ICME), IEEE, 2023, pp. 1601--1606.

\bibitem{hu2018squeeze}
J.~Hu, L.~Shen, G.~Sun, Squeeze-and-excitation networks, in: Proceedings of the IEEE conference on computer vision and pattern recognition, 2018, pp. 7132--7141.

\bibitem{simonyan2014very}
K.~Simonyan, A.~Zisserman, Very deep convolutional networks for large-scale image recognition, arXiv preprint arXiv:1409.1556 (2014).

\bibitem{zhao2016loss}
H.~Zhao, O.~Gallo, I.~Frosio, J.~Kautz, Loss functions for image restoration with neural networks, IEEE Transactions on computational imaging 3~(1) (2016) 47--57.

\bibitem{bergmann2018improving}
P.~Bergmann, S.~L{\"o}we, M.~Fauser, D.~Sattlegger, C.~Steger, Improving unsupervised defect segmentation by applying structural similarity to autoencoders, arXiv preprint arXiv:1807.02011 (2018).

\bibitem{dong2020multi}
H.~Dong, J.~Pan, L.~Xiang, Z.~Hu, X.~Zhang, F.~Wang, M.-H. Yang, Multi-scale boosted dehazing network with dense feature fusion, in: Proceedings of the IEEE/CVF conference on computer vision and pattern recognition, 2020, pp. 2157--2167.

\bibitem{ren2018gated}
W.~Ren, L.~Ma, J.~Zhang, J.~Pan, X.~Cao, W.~Liu, M.-H. Yang, Gated fusion network for single image dehazing, in: Proceedings of the IEEE conference on computer vision and pattern recognition, 2018, pp. 3253--3261.

\bibitem{dong2020physics}
J.~Dong, J.~Pan, Physics-based feature dehazing networks, in: Computer Vision--ECCV 2020: 16th European Conference, Glasgow, UK, August 23--28, 2020, Proceedings, Part XXX 16, Springer, 2020, pp. 188--204.

\bibitem{guo2023scanet}
Y.~Guo, Y.~Gao, W.~Liu, Y.~Lu, J.~Qu, S.~He, W.~Ren, Scanet: Self-paced semi-curricular attention network for non-homogeneous image dehazing, in: Proceedings of the IEEE/CVF Conference on Computer Vision and Pattern Recognition, 2023, pp. 1884--1893.

\bibitem{zhang2024waveletformernet}
S.~Zhang, Z.~Tao, S.~Lin, Waveletformernet: A transformer-based wavelet network for real-world non-homogeneous and dense fog removal, Image and Vision Computing 146 (2024) 105014.

\bibitem{jin2022structure}
Y.~Jin, W.~Yan, W.~Yang, R.~T. Tan, Structure representation network and uncertainty feedback learning for dense non-uniform fog removal, in: Asian Conference on Computer Vision, Springer, 2022, pp. 155--172.

\bibitem{park2020contrastive}
T.~Park, A.~A. Efros, R.~Zhang, J.-Y. Zhu, Contrastive learning for unpaired image-to-image translation, in: Computer Vision--ECCV 2020: 16th European Conference, Glasgow, UK, August 23--28, 2020, Proceedings, Part IX 16, Springer, 2020, pp. 319--345.

\end{thebibliography}





\end{document}